\documentclass[10pt, conference, compsocconf]{IEEEtran}

\usepackage{cite}
\usepackage{amsmath,amssymb,amsfonts}
\usepackage{algorithmic}
\usepackage{graphicx}
\usepackage{textcomp}
\usepackage{xcolor}
\def\BibTeX{{\rm B\kern-.05em{\sc i\kern-.025em b}\kern-.08em
    T\kern-.1667em\lower.7ex\hbox{E}\kern-.125emX}}
\usepackage[algo2e,ruled,linesnumbered,noend]{algorithm2e}
\usepackage{multirow}
\usepackage{url}

\newcommand{\indep}{\perp \!\!\! \perp}
\newcommand{\ncd}{\textsc{ncd}}
\newcommand{\ncda}{\textsc{ncda}}
\newcommand{\gencda}{\textsc{ge\ncda{}}}

\newcommand{\gcm}{\textsc{gcm}}
\newcommand{\tvae}{\textsc{tvae}}
\newcommand{\ctgan}{\textsc{ctgan}}
\newcommand{\rnd}{\textsc{rnd}}

\newcommand{\abalone}{\texttt{abalone}}
\newcommand{\old}{\texttt{oldf}}
\newcommand{\dwd}{\texttt{dwd}}
\newcommand{\undata}{\texttt{undata}}
\newcommand{\diabets}{\texttt{diabets}}

\begin{document}
\bstctlcite{IEEEexample:BSTcontrol}

\title{Boosting Synthetic Data Generation with Effective Nonlinear Causal Discovery}

\author{\IEEEauthorblockN{Martina Cinquini}
\IEEEauthorblockA{\textit{University of Pisa}\\
Pisa, Italy \\
martina.cinquini@phd.unipi.it}
\and
\IEEEauthorblockN{Fosca Giannotti}
\IEEEauthorblockA{\textit{ISTI-CNR}\\
Pisa, Italy \\
fosca.giannotti@isti.cnr.it}
\and
\IEEEauthorblockN{Riccardo Guidotti}
\IEEEauthorblockA{\textit{University of Pisa}\\
Pisa, Italy \\
riccardo.guidotti@unipi.it}
}


\newif\ifdraft
\drafttrue
\newcommand{\MC}[1]{\ifdraft{{\color{red} [MC]: {#1}}}\else{\vspace{0ex}}\fi}


\maketitle

\begin{abstract}
Synthetic data generation has been widely adopted in software testing, data privacy, imbalanced learning, artificial intelligence explanation, etc. 
In all such contexts, it is important to generate plausible data samples.
A common assumption of approaches widely used for data generation is the independence of the features.
However, typically, the variables of a dataset depend on one another, and these dependencies are not considered in data generation leading to the creation of implausible records. 
The main problem is that dependencies among variables are typically unknown.
In this paper, we design a synthetic dataset generator for tabular data that is able to discover nonlinear causalities among the variables and use them at generation time.
State-of-the-art methods for nonlinear causal discovery are typically inefficient.
We boost them by restricting the causal discovery among the features appearing in the frequent patterns efficiently retrieved by a pattern mining algorithm.
To validate our proposal, we design a framework for generating synthetic datasets with known causalities.
Wide experimentation on many synthetic datasets and real datasets with known causalities shows the effectiveness of the proposed method.
\end{abstract}

\begin{IEEEkeywords}
Data Generation, Causal Discovery, Pattern Mining, Synthetic Datasets, Explainability
\end{IEEEkeywords}

\section{Introduction}
In many real-world applications, it is fundamental to rely on synthetic data, especially when real data can be difficult to obtain due to privacy issues, temporal or budget constraints, or the unavailability of large quantities.
Synthetic data are used for validating data discovery applications and for testing software in a controlled environment that satisfies specific conditions~\cite{jeske2006synthetic,michael1997genetic}. 
In machine learning, synthetic data are increasingly being used for addressing imbalanced learning~\cite{chawla2002smote}, for training a model with the intention of transfer learning to real data~\cite{pan2010survey}, or, in the last days, for providing explanations of obscure decision systems~\cite{ribeiro2016should}. 
Indeed, various studies show the benefits of using synthetic data located in the neighborhood of available real instances for learning predictive models~\cite{yeung2007localized,ng2007image}, or for explaining the reasons for the prediction~\cite{guidotti2019survey}.

In these scenarios, the methods used for synthetic data generation are either simple but efficient random approaches assuming uniform distribution for all the variables~\cite{ribeiro2016should}, or complex and time expensive methods such as Generative Adversarial Networks (GAN)~\cite{goodfellow2014generative}.  
However, typically, only a few of them rely on explicit knowledge about possible linear and/or nonlinear dependencies among the variables.
In particular, common generative approaches work on the assumption that the variables of the dataset to generate are independent.
Such an assumption does not guarantee a reliable synthetic generation of the dataset under analysis.
On the other hand, GAN-like approaches can theoretically learn possible dependencies, but these are not explicitly represented, and there is no guarantee that they are followed in the data generation process. 

Therefore, a crucial problem in synthetic data generation is that dependencies among variables are not used because they are typically unknown.
Our idea is to design a technique for synthetic data generation that accounts for dependencies among variables by exploiting a \textit{causal discovery} algorithm.
Causal discovery algorithms take as input a set of variables belonging to a dataset and determine which are the causal relationships among them~\cite{mooij2016distinguishing,guo2020survey}. 
In particular, we focus on \textit{nonlinear causal discovery}~\cite{hoyer2008nonlinear}.
If the dataset under analysis is continuous-valued, methods based on linear causal models are commonly applied~\cite{pearl2000causality}.
This typically happens because linear models are well understood and not necessarily because the true causal relationships are believed to be linear~\cite{hoyer2008nonlinear}.
However, in reality, many causal relationships are nonlinear, raising doubts on the reliability and usability of linear methods.
In addition, in~\cite{hoyer2008nonlinear} it is shown that considering nonlinear causal relationships plays a primary role in the identification of causal directions.
For these reasons, we start from the \textsc{n}onlinear \textsc{c}ausal \textsc{d}iscovery method described in~\cite{hoyer2008nonlinear} (\ncd{}) to design our synthetic data generator exploiting causal relationships.
Besides non-linearity, the \ncd{} is able to consider and discover not only binary relationships but also multivariate ones.
Unfortunately, the \ncd{} approach is inefficient and can only be employed to reveal the causal relationships of datasets with a very small number of variables.
Thus, it is not practically usable for applications on real datasets.
Our proposal is to boost \ncd{} by restricting the search of causalities among the features appearing in the patterns returned by a pattern mining algorithm executed on the same dataset.

Our contribution is twofold. 
First, we design an efficient method for nonlinear causal discovery based on pattern mining named \ncda{} (\textsc{n}onlinear \textsc{c}ausal \textsc{d}iscovery with \textsc{A}priori).
Second, we implement \gencda{}, a \textsc{ge}nerative method based on \ncda{}.
Moreover, to validate our proposal, we realized a framework for generating synthetic datasets with known causalities.
We report wide experimentation on synthetic datasets and real datasets with known causalities highlighting the effectiveness of our proposals both in terms of time and accuracy for causal discovery and data generation.

The rest of the paper is organized as follows. 
Section~\ref{sec:related} discusses related works.
Section~\ref{sec:background} recalls the notions needed to understand the proposed methods which are illustrated in Section~\ref{sec:method}.
Section~\ref{sec:experiments} presents the experimental results. 
Finally, Section~\ref{sec:conclusion} concludes the paper by discussing known limitations and proposing future research directions.

\section{Related Works}
\label{sec:related}
In this section, we review existing proposals in the literature related to causal discovery and synthetic data generation.

The discovery of causal relationships between a set of observed variables is a fundamental problem in science because it enables predictions of the consequences of actions~\cite{pearl2000causality}.
Thus, the development of automatic and data-driven causal discovery methods constitutes an important research topic~\cite{pearl2000causality,spirtes2000causation,spirtes2016causal}.
A standard approach for causal discovery is to estimate a Markov equivalence class of directed acyclic graphs from the data~\cite{pearl2000causality,spirtes2000causation}. 
The independence tests often adopt linear models with Gaussian noise~\cite{spirtes2000causation}. 
However,~\cite{shimizu2006linear} shows that non-Gaussian noise in linear models can actually help in distinguishing the causal directions. 
In~\cite{hoyer2008nonlinear,chen2013nonlinear,cai2019causal} is shown that nonlinear models can play a role similar to that of non-Gaussianity.
Indeed, when causal relationships are nonlinear it allows the identification of causal directions by breaking the symmetry between the variables. 
Also,~\cite{friedman2000gaussian} shows that non-invertible functional relationships between the variables can provide clues to identify causal relationships. 
For nonlinear models with additive noise almost any nonlinear relationship (invertible or not) typically suggests identifiable models. 
In~\cite{chen2013nonlinear} is presented a nonlinear causal discovery approach for high dimensional data based on the idea of mapping the observations to high dimensional space with a kernel such that the nonlinear relations become simple linear ones.
A problem of the aforementioned nonlinear causal discovery approaches is that they can miss detecting indirect causal relationships, which are frequently encountered in practice and result from omitted intermediate causal variables. 
In~\cite{cai2019causal} is proposed a cascade nonlinear additive noise model to represent such causal influences in a way that each direct causal relation follows the nonlinear additive noise model only observing initial cause and final effect.
Despite various advantages of the recent proposals, to the best of our knowledge, the nonlinear causal discovery described in~\cite{hoyer2008nonlinear} is the only one that allows managing not only binary relationships.
For this reason, we adopt it as starting point of our procedure.
However, we highlight that our proposal is a framework that can exploit the preferred causal discovery method.

The need to generate synthetic data derives from the first data imputation work to solve the problem of non-responses in statistical surveys~\cite{dandekar2018comparative}.
In~\cite{rubin2004multiple} is described one of the first multiple imputation techniques used to synthetically generate the values of a set of missing attributes for the records of the dataset.
In~\cite{raghunathan2003multiple} are implemented and extend multiple imputation approaches for the specific case of synthetic data generation.
In machine learning, synthetic data are often used for handling the classification task in case of imbalanced data and for addressing the problem of outcome explanation of black-box classifiers.
Concerning imbalanced learning, the {\scshape smote} algorithm~\cite{chawla2002smote} generates an arbitrary number of synthetic instances to shift the learning bias toward the minority class. 
The {\scshape adasyn} approach~\cite{he2008adasyn} is based on the idea of adaptively generating minority data samples according to their distributions. 
In practice, the method generates more synthetic samples for minority classes that are harder to learn compared to those minority samples that are easier to learn.
In~\cite{reiter2005using} is introduced an alternative method to synthesize data through a non-parametric technique that uses classification and regression trees.
In~\cite{ping2017datasynthesizer} is proposed \textit{DataSynthesizer}, a technique that captures the underlying correlation structure between different attributes by building a Bayesian network.
Another recent generator is the \textit{Synthetic Data Vault} (\textsc{sdv})~\cite{patki2016synthetic} which uses the multivariate Gaussian copula (\gcm{}) to calculate the covariances between the input columns. 
After that, the distributions and covariances are sampled to return synthetic data.
In addition, recently, many generative models have been developed based on Generative Adversarial Networks (GANs) and their extensions~\cite{goodfellow2014generative} and autoencoders~\cite{makhzani2015adversarial}. 
Their success is due to their high effectiveness and flexibility in generating and representing data.
These approaches are particularly employed for the generation of synthetic images and, in general, for unstructured data such as text or time series.
However, recently, they are being effectively applied also for the generation of tabular data.
In~\cite{xu2018synthesizing} and~\cite{xu2019modeling} are proposed synthetic tabular data generators using GANs, Conditional GANs and Variational Autoencoders (VAE). 
Unfortunately, since they are based on deep learning procedures, they require a non-negligible amount of data and a considerable amount of time.
Finally, in eXplainable Artificial Intelligence (XAI)~\cite{guidotti2019survey}, data generation approaches are used to learn interpretable models able to mimic black-box decision systems. 
{\scshape lime}~\cite{ribeiro2016should} explains the local behavior of a black-box classifier by learning a linear model on synthetic data generated around the instance to explain using a \textit{normal} distribution.
{\scshape lore}~\cite{guidotti2019factual} is another local explanation method that exploits a synthetic neighborhood generation based on a genetic algorithm to create a more compact dataset around the explained instance.

Among wide literature concerning synthetic data generation and causality, integrated approaches are a recently challenging research area. In~\cite{runge2019inferring}, authors
implement \textit{CauseMe}, a platform to benchmark causal discovery methods acting on time series. 
In 2021, Lawrence et al.~\cite{lawrence2021data} propose an easily parameterizable process that provides the capability to generate synthetic time series from vastly different scenarios. 
Lastly, Wood-Doughty et al.~\cite{woodfoughty2021generating} present a synthetic text generator to evaluate causal inference (not causal discovery) methods. 
Thus, to the best of my knowledge, no state-of-the-art synthetic data generators explicitly allow for encoding causal relationships.


\section{Setting the Stage}
\label{sec:background}
In this paper, we address the problem of \textit{synthetic data generation with unknown causal dependencies}.
Consider a dataset $X = \{x_1, x_2, \dots, x_n\}$ formed by a set of $n$ instances such that each instance $x_i \in \mathbb{R}^{m}$ consists of $m$ values.
We adopt $x_i$ to indicate the $i$-th row of $X$, i.e., the $i$-th instance, while we use $x^{(j)}$ to indicate the $j$-th column of $X$.
We use the notation $a^{(j)}$ to indicate the attribute name of the $j$-th feature, and $v_i^{(j)}$ to refer to the value belonging to the domain of $a^{(j)}$ of the $i$-th instance.
E.g., $a^{(j)} = \mathit{age}$ and $v_i^{(j)} = 32$.
Thus, $x^{(j)} = [v^{(j)}_1, \dots, v^{(j)}_n]$.
Given a dataset $X$ we assume that exist some unknown causal dependencies among the $m$ variables of $X$.
We model the dependencies with a Directed Acyclic Graph (DAG) $G$: every node models a feature (variable), and there is a directed edge from $i$ to $j$ if $i$ contributes in causing $j$~\cite{hoyer2008nonlinear}.
Given $X$ having unknown causal dependencies $G$, our objective is \emph{(i)} to \textit{discover} the causal dependencies of $X$, named $\tilde{G}$, and then, \emph{(ii)} to \textit{generate} a synthetic version of $X$, named $\tilde{X}$, respecting the discovered causal dependencies $\tilde{G}$.
The goals are \emph{(i)} to accurately discover the dependencies such that the differences between the real unknown DAG $G$ and discovered DAG $\tilde{G}$ are minimized, and \emph{(ii)} to generate $\tilde{X}$ such that some interest properties \mbox{that are valid for $X$ hold also for $\tilde{X}$.}

\smallskip
We keep our paper self-contained by summarizing here the key concepts necessary to comprehend our proposal.

\subsection{Nonlinear Causal Discovery}
\label{sec:hoyer}
Given a dataset $X$, the objective of causal discovery is to infer as much as possible about the mechanism generating the data.
In particular, the goal is to discover the graph $G$ modeling the dependencies among variables.

In~\cite{hoyer2008nonlinear} is described the \textsc{n}onlinear \textsc{c}ausal \textsc{d}iscovery (\ncd{}) approach that we adopt as starting point for our proposal.
Hoyer et al. adopt the following assumptions.
Given a DAG $G$ describing the causal relationships of a dataset $X$, each feature $x^{(j)}$ is associated with a node $j$ in $G$, and the values of $x^{(j)}$ are obtained as a function of its parents in $G$, plus some independent additive noise $\nu^{(j)}$, i.e., 
\begin{equation}
\label{eq:hoyer}
    x^{(j)} = f_j(\mathit{pa}(j)) + \nu^{(j)}
\end{equation}
where $f_j$ is an arbitrary function (possibly different for each $j$), $\mathit{pa}(j)$ is a vector containing the elements $x^{(j)}$ such that there is an edge from $i$ to $j$ in $G$, i.e., $\mathit{pa}(j)$ returns the parents of $j$.
The noise variables $\nu^{(j)}$ may have arbitrary probability densities $p_{\nu_j}(\nu_j)$ and are independent from $\mathit{pa}(j)$, i.e., $\nu_j \indep \mathit{pa}(j)$.
\ncd{} includes the special case when all the $f_j$ are linear and all the $p_{\nu_j}$ are Gaussian, yielding the standard linear–Gaussian model family~\cite{spirtes2000causation}. 
Also, when the functions are linear but the densities are non-Gaussian it reduces to linear–non-Gaussian models~\cite{shimizu2006linear}.

The \ncd{} method works as follows.
Given a dataset $X$, it selects any possible (nonempty) subsets of features $U = \{a^{(j_1)}, \dots, a^{(j_k)}\}$ and $V = \{a^{(j'_1)}, \dots, a^{(j'_{k'})}\}$ (with $U \cap V = \emptyset$) and repeats the following procedure.
First, it tests whether $U$ and $V$ are statistically independent. 
If they are not, it continues as in the following.
It verifies if Equation~\ref{eq:hoyer} is consistent with the data by making a nonlinear regression of $V$ on $U$, i.e., $V = f(U) + \nu$, to obtain an estimation $\hat{f}$ of $f$.
Then it calculates the residuals $\hat{\nu} = V - \hat{f}(U)$, and tests if $\hat{\nu}$ is independent from $U$.
If this condition is verified, i.e., $\hat{\nu} \indep U$, then the model of Equation~\ref{eq:hoyer} is accepted, otherwise it is rejected.
The same procedure is applied to test if the reversed model fits the data, i.e., $U = f(V) + \nu$, to check if $\hat{\nu} \indep V$.

The aforementioned procedure can have five possible outcomes.
First, $U$ and $V$ are statistically independent and the procedure is not applied.
Second, if $\hat{\nu}$ is independent from $U$ and dependent from $V$, i.e., $\hat{\nu} \indep U \wedge \nu \not\indep V$, then we deduce that $U$ causes $V$ ($U \rightarrow V$).
Third, if $\hat{\nu} \indep V \wedge \hat{\nu} \not\indep U$,
we deduce that $V$ causes $U$ ($V \rightarrow U$).
Fourth, if 
$\hat{\nu} \not\indep U \wedge \hat{\nu} \not\indep V$, neither direction is consistent with the dataset and we cannot deduce anything.
Fifth, if 
$\hat{\nu} \indep U \wedge \hat{\nu} \indep V$, both models are accepted and we cannot deduce any model from the dataset.

The selection of a particular independence test or nonlinear regressor is not constrained to specific implementations.
In particular, in~\cite{hoyer2008nonlinear} the authors adopt the Hilbert-Schmidt Independence Criterion (HSIC) as independence test~\cite{gretton2005kernel}, and Gaussian Processes for nonlinear regressions~\cite{rasmussen2006gaussian}.

The \ncd{} method can be used for checking binary causal relationships, i.e., when $|U|=|V|=1$, but can also be used for an arbitrary number of observed variables.
However, as stated in~\cite{hoyer2008nonlinear}, is feasible only for datasets with a low number of features ($m \leq 7$).
For this reason, we propose a ``filtering'' approach based on frequent pattern mining that reduces the total number of relationships to be tested by \ncd{}.

\subsection{Pattern Mining and Apriori}
\label{sec:apriori}
Pattern mining methods allow to discover interesting patterns describing relationships between features in the data in an efficient manner~\cite{tan2005introduction}.
The relationships that are hidden in the data can be expressed as a collection of \textit{frequent itemsets}.

Let $T = \{t_1, \dots, t_n \}$ be a set of $n$ transactions (or baskets) and $E = \{i_1, \dots, i_m \}$ a set of $m$ items, a basket $t_i$ is a subset of items such that $\emptyset \subset t_i \subseteq E$.
A set of items that are \textit{frequent} in $T$ is called \emph{itemset} or \emph{pattern}.
An itemset $S$ is frequent if its \textit{support} is higher than a $\mathit{min\_sup}$ parameter.
The support over $T$ of an itemset $S$ is defined as $supp_T(S) = |\{t_i \in T | S \subseteq t_i\}| / |T|$.
The problem of finding the \emph{frequent itemsets} from a dataset of transactions $T$ requires to find in a set of transactions all the itemsets having support greater or equal than $\mathit{min\_sup}$. 
The search space of itemsets that need to be explored to find the frequent itemsets is exponentially large ($2^m-1$).
Indeed, the set of all possible itemsets forms a lattice structure 
and using a brute force algorithm makes the problem intractable for large datasets. 

\emph{Apriori} is the most famous algorithm for finding frequent itemsets~\cite{agrawal1994fast}.
Apriori proposes an effective way to eliminate candidate itemsets without counting their support. 
It is based on the principle that if an itemset is frequent, then all of its subsets must also be frequent.
This principle is used for pruning candidates during the itemset generation. 

Our idea is to exploit Apriori to check for the presence of causal relationships only for features for which exists a frequent pattern containing that feature.
In particular, we will focus on maximal itemsets.
A frequent itemset is \textit{maximal} if there is no other frequent itemset containing it.


\section{Data Generation with Causal Knowledge}
\label{sec:method}
In this section, first we describe our idea for boosting the nonlinear causal discovery algorithm proposed in~\cite{hoyer2008nonlinear} with pattern mining and making it practically usable on real multivariate datasets.
Then we describe the data generative process that takes as input the causal relationships discovered and returns a synthetic dataset that respects them.

\subsection{Pattern Mining-based Nonlinear Causal Discovery}
\label{sec:ncda}
In Section~\ref{sec:hoyer} we have described the method of causal discovery based on nonlinear models with additional noise (\ncd{}). 
We have shown that: \textit{(i)} the procedure allows for unambiguous identification of the causal relationships, and that \textit{(ii)} unlike other causal discovery methods, \ncd{} is also applicable to multivariate data.
Despite these advantages, the main problem with \ncd{} is the need to explore all the possible direct acyclic graphs (DAGs) to identify the final causal structure $\tilde{G}$. 
It follows that the computational complexity of the algorithm is super-exponential~\cite{robinson1977counting}. 
Hence, we propose a solution to this bottleneck by exploiting the Apriori algorithm. 

Our idea can be summarized as follows.
First, we apply Apriori to the dataset under study for extracting the frequent patterns. 
Then, we test the causal relationships considering only the combination of variables appearing together in any of the itemset extracted with Apriori.
In other words, we use Apriori as a filter to reduce the number of possible combinations and to reduce the search space for \ncd{}.
It follows that the \ncd{} approach is no longer applied on all the possible combinations of variables in the dataset, but only on those for which there are \mbox{frequent patterns that highlight a correlation.}

Our intuition comes from the fact that, thanks to the extraction of frequent itemsets, Apriori provides useful information on the correlations among the variables. 
The fact that variables are correlated does not necessarily indicate a causal relationship. 
Indeed, while \textit{causation} and \textit{correlation} can exist at the same time, correlation does not necessarily imply causation.
Correlation means there is a relationship or pattern among certain variables, while causation means that one (set of) variable(s) causes another one to occur.
However, there is a need for some ``link'' between the variables involved for causality to exist. 
In other terms, we exploit the presence of variables in a pattern and their observed correlation as a ``clue'' about the possible presence of a causal relationship.
This assumption is the core of our intuition, and it suggested introducing an intermediate filtering step in the discovery of the causal structure by exploiting pattern mining.

\begin{algorithm2e}[t]
	\footnotesize
    \caption{\mbox{\ncda{}($X$, $\mathit{n\_bins}$, $\mathit{min\_sup}$, $\mathit{max\_len}$, $\alpha$)}}
	\label{alg:ncda}
	\SetKwInOut{Input}{Input}
	\SetKwInOut{Output}{Output}
	\Input{$X$ - dataset,
	    $\mathit{n\_bins}$ - nbr of bins,
	    $\mathit{min\_sup}$ - min supp.,\\
	    $\mathit{max\_len}$ - max length,
	    $\alpha$ - p-value thr.
	}
	\Output{$\tilde{G}$ - DAG modeling causal relationships
	}
	
	\BlankLine
	$\tilde{G} \leftarrow \emptyset$;\tcp*[f]{\texttt{\scriptsize init. empty DAG}}\\
	$T \leftarrow \mathit{discretize}(X, \mathit{n\_bins})$;\tcp*[f]{\texttt{\scriptsize cont. to cate.}}\\
	$\mathcal{S} \leftarrow \textsc{apriori}(T, \mathit{min\_sup}, \mathit{max\_len})$;\tcp*[f]{\texttt{\scriptsize run Apriori}}\\
	\For{$S \in \mathcal{S}$}{
	    $V \leftarrow \mathit{getVariables}(S)$; \tcp*[f]{\texttt{\scriptsize extract variables}}\\
	    $C \leftarrow \ncd{}(X^{(V)}, \alpha)$; \tcp*[f]{\texttt{\scriptsize run NCD}}\\
	    $\tilde{G} \leftarrow \mathit{updateGraph}(\tilde{G}, C)$; \tcp*[f]{\texttt{\scriptsize update graph}}\\
	}
	
    \Return{$\tilde{G}$}\;
\end{algorithm2e}

We name our proposal \ncda{} (nonlinear causal discovery with Apriori) and we present it in Algorithm~\ref{alg:ncda}.
\ncda{} takes as input a dataset $X$ formed by continuous variables and returns the DAG $\tilde{G}$ that describes the causal structure of $X$.
First, it initializes an empty DAG $\tilde{G}$ (line 1).
Then, since \ncd{} works on continuous variables, while \textsc{apriori} works on transactional data, we have to turn the dataset $X$ into its transactional version $T$.
\ncda{} implements this step with the $\mathit{discretize}$ function that works as follows.
For each feature $j$, \ncda{} divides the set of values $x^{(j)}$ into $\mathit{n\_bins}$ equal sized bins.
For instance, if $a^{(j)}$ is describing the \textit{age} that in $X$ ranges from $20$ to $80$ and $\mathit{n\_bins} = 5$, then the $j$-th feature will be described with $5$ categorical values each one representing $12$ values, i.e., $\mathit{age}\_[20, 32], \mathit{age}\_[32,44], \mathit{age}\_[44, 56], \mathit{age}\_[56, 68], \mathit{age}\_[68,$ $80]$. 
Thus, a record $x_i = \{(\mathit{age}, 30), (\mathit{insulin}, 94), (\mathit{BMI}, $ $25.3)\}$ is translated into the transaction $t_i = \{\mathit{age}\_[20, 32],$ $ \mathit{insulin}\_[90, 110], \mathit{BMI}\_[20.5, 26.8]\}$.
After that, \ncda{} applies \textsc{apriori} on $T$ using the parameters $\mathit{min\_sup}$ and $\mathit{max\_len}$ regulating the minimum support and maximum pattern length, respectively (line 3).
The set of maximal itemsets\footnote{We use \textit{maximal} itemsets because \ncd{} tests all the possible combinations of the input variables, therefore it would not have been useful to test also the variables of itemsets which are subsets of the maximal ones.} is stored into $\mathcal{S}$.
An example of itemset $S \in \mathcal{S}$ can be $S = \{\mathit{age}\_[20, 32], \mathit{BMI}\_[20.5, 26.8]\}$, meaning that there is a high number of co-occurrences of records in $T$ with \textit{age} in 20-32 and \textit{BMI} in 20.5-26.8.
This is the ``pattern mining clue'' that \ncda{} exploits to check if there is a causal relationship between \textit{age} and \textit{BMI}.

For each maximal itemset $S \in \mathcal{S}$ (lines 4--7), \ncda{} repeats the following steps.
First, it extracts from the itemset $S$ the variables $V$ present (line 5).
In our example, from the pattern $S = \{\mathit{age}\_[20,32], \mathit{BMI}\_[20.5, 26.8]\}$ we obtain the features $V = \{\mathit{age}, \mathit{BMI}\}$.
Then, it runs the \ncd{} method on the dataset $X$ considering only the variables in $V$, i.e. $X^{(V)}$ (line 6).
This step is where the Apriori filter acts: \ncd{} tests all the possible DAGs among those that can be derived from the features in $V$.
We underline that, instead of testing all the possible DAGs from the $m$ features of $X$ as proposed in~\cite{hoyer2008nonlinear}, \ncd{} only tests the possible DAGs from the $|V|$ features of $V$ with $2 \leq |V| \leq \mathit{max\_len} \ll m$.
Obviously, more than one pattern can suggest checking causations for the same set of variables $V$.
In this case, \ncd{} is executed only the first time that a specific set of variables $V$ is analyzed.
The $\alpha$ parameter is the p-value threshold used for the HSIC independence test.
Finally, if there are causal relationships $C$ identified by \ncd{}, i.e., $C \neq \emptyset$, \ncda{} updates the DAG $\tilde{G}$ by adding the corresponding edges to model these relationships (line 7).
We highlight that \ncd{} returns all the causal relationships consistent with the data $X^{(V)}$, including possible sub-relationships. 
To the aim of returning only the most representative ones, we consider only the causal relationships $C$ returned by \ncd{} with the highest average level of p-values among the various detected dependencies\footnote{With p-value we mean the probability of obtaining a result of the statistical test at least as extreme as the one actually observed assuming that the null hypothesis is true, i.e., the variables considered are independent.}.
The above heuristic is also used to guarantee that the DAG returned is a valid one.

\begin{algorithm2e}[t]
	\footnotesize
    \caption{\gencda{}($X$,$\tilde{G}$,$\mathcal{F}$)}
	\label{alg:gencda}
	\SetKwInOut{Input}{Input}
	\SetKwInOut{Output}{Output}
	\Input{$X$ - real dataset,
	    $\tilde{G}$ - DAG modeling causal relationships,
	    $D$ - set of distributions
	}
	\Output{$\tilde{X}$ - synthetic dataset
	}
	
	\BlankLine
	$\tilde{X} \leftarrow \emptyset$;\tcp*[f]{\texttt{\scriptsize init. empty dataset}}\\
	$\tilde{G}' \leftarrow \mathit{sort}(\tilde{G})$;\tcp*[f]{\texttt{\scriptsize topological sort}}\\
	\For(\tcp*[f]{\texttt{\scriptsize for each node/variable j}}){$j \in \tilde{G}'$}{
    	\If(\tcp*[f]{\texttt{\scriptsize node j has no parents}}){$\mathit{pa}(j) = \emptyset$}{
        	$d \leftarrow \mathit{fit}(X^{(j)}, D)$; \tcp*[f]{\texttt{\scriptsize fit distribution}}\\
        	$\tilde{X}^{(j)} \leftarrow \mathit{sample}(d)$; \tcp*[f]{\texttt{\scriptsize sample from distrib.}}\\
    	}\Else{
        	$r \leftarrow \mathit{train}(X^{\mathit{pa}(j)}, X^{(j)})$; \tcp*[f]{\texttt{\scriptsize train regressor}}\\
        	$\tilde{X}^{(j)} \leftarrow \mathit{apply}(r, \tilde{X}^{\mathit{pa}(j)})$; \tcp*[f]{\texttt{\scriptsize apply regressor}}\\
    	}
	}
    \Return{$\tilde{X}$}\;
\end{algorithm2e}

\subsection{Causality-based Synthetic Data Generator}
\label{sec:gencda}
In this section we present \gencda{}, a synthetic data {\scshape ge}nerator based on \ncda{}.
\gencda{} exploits the causal relationships discovered by \ncda{} to generate a synthetic dataset respecting such causal structure.
The pseudo-code of \gencda{} is reported in Algorithm~\ref{alg:gencda}.

\gencda{} takes as input the \textit{real} dataset $X$ that has to be extended with \textit{synthetic} data, the DAG $\tilde{G}$ extracted from $X$ by \ncda{} and a set of distributions to test $D$.
First, \gencda{} initializes an empty synthetic dataset $\tilde{X}$ and applies a topological sorting\footnote{A topological sorting for a DAG is a linear ordering of vertices such that for every directed edge from $i$ to $j$, vertex $i$ comes before $j$ in the ordering.} on $\tilde{G}$ (lines 1 and 2).
The topological sorting allows \gencda{} to consider first the independent variables and then the dependent ones.
In this way, when it is time to generate the dependent variables, the independent ones involved in the causal relationships have been already generated and can be actively used.
Then, according to the topological ordering in $\tilde{G}'$ it repeats the steps in lines 3--9.
Given the vertex $j$, and therefore the corresponding $j$-th variable in $X$, if the set of parents $\mathit{pa}(j)$ for $j$ is empty (line 4), then the variable is independent, otherwise it is a dependent one.
If $j$ is an independent variable, \gencda{} tries to identify the best distribution $d \in D$ that fits with the data in $X^{(j)}$ using the Kolmogorov-Smirnov test (line 5).
After that, it samples from the distribution $d$ and synthetically generates the values $\tilde{X}^{(j)}$ for the $j$-th variable (line 6).
On the other hand, if $j$ is a dependent variable, then \gencda{} learns a regressor model $r$ on the features $X^{\mathit{pa}(j)}$ for predicting $X^{(j)}$ (line 8).
Then, it applies the regressor $r$ on the data $\tilde{X}^{\mathit{pa}(j)}$ generated in the previous iterations and synthetically creates the values $\tilde{X}^{(j)}$ for the $j$-th variable respecting the causal relationships with their parents (line 9).

The function $\mathit{fit}$ in line 5 of \gencda{} returns the distribution $d$ among those in $D$ that minimizes the Sum of Squared Error (SSE) between the probability density of the distribution and the estimate of that of the data.
For the functions $\mathit{train}$ and $\mathit{apply}$ in lines 8 and 9 of \gencda{} we exploit an ensemble of four different regressors: Gaussian Process Regressor (GPR), Support Vector Machine (SVM), k-Nearest Neighbors (kNN), and Decision Tree Regressor (DTR).
The predicted value used as a dependent variable for the synthetic dataset is the mean of the predictions of the four regressors.

\section{Experiments}
\label{sec:experiments}
In this section, we show the impact of Apriori on the performance of \ncd{}\footnote{
    Python code and datasets available at: \url{https://github.com/marti5ini/GENCDA}.
    Experiments were run on MacBook Pro, Apple M1 3.2 GHz CPU, 8 GB LPDDR4 RAM.
}.
First, we illustrate the evaluation measures adopted.
Then, we detail the framework developed for generating synthetic datasets with known causalities, and we describe the real datasets.
After that, we show the baselines used to compare with our proposal. 
Finally, we report the experimental settings, the empirical evaluation, and the sensitivity analysis.

\begin{figure*}[t]
    \centering
    \includegraphics[trim=10mm 10mm 10mm 5mm, clip, width=0.2\linewidth]{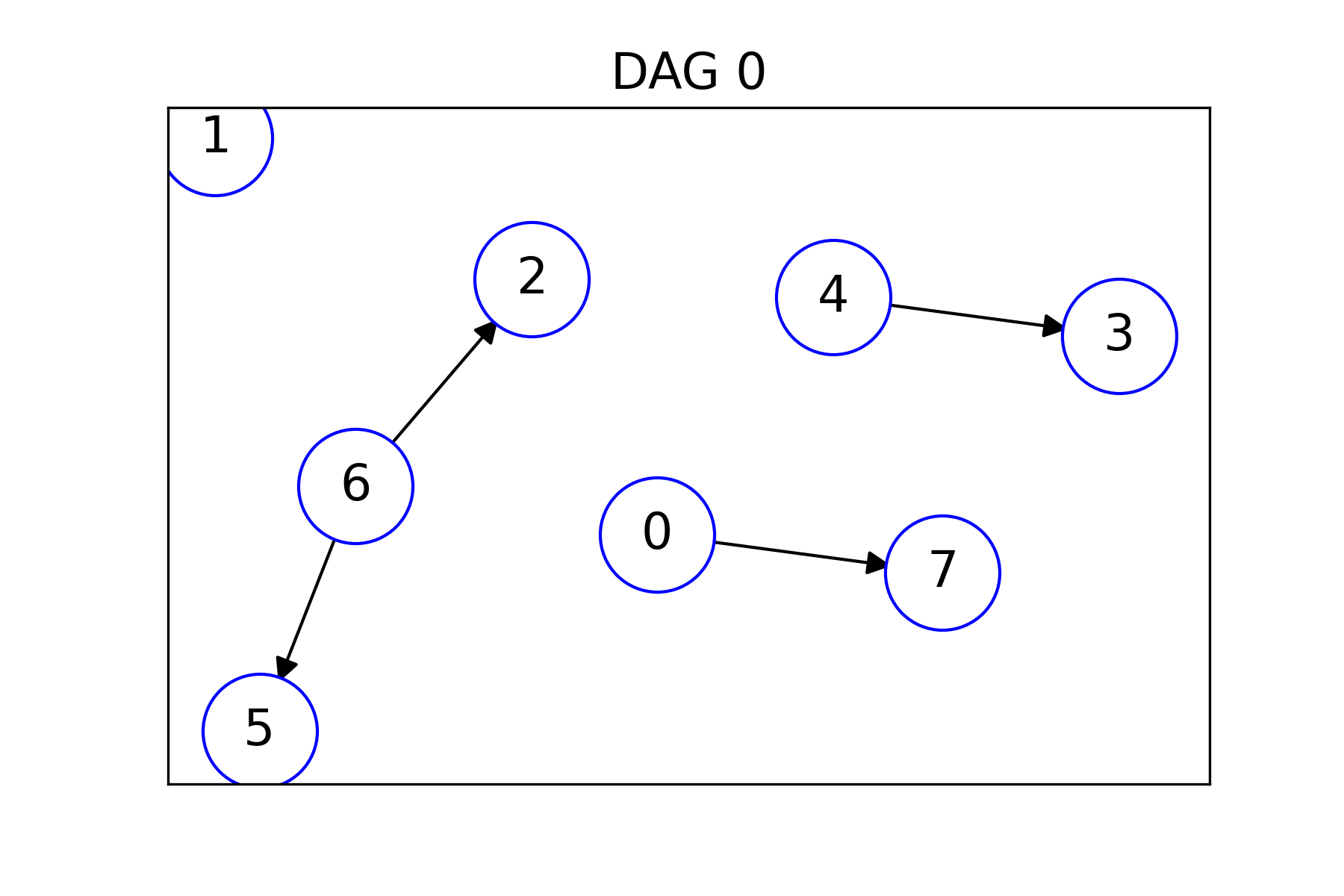}%
    \includegraphics[trim=10mm 10mm 10mm 5mm, clip, width=0.2\linewidth]{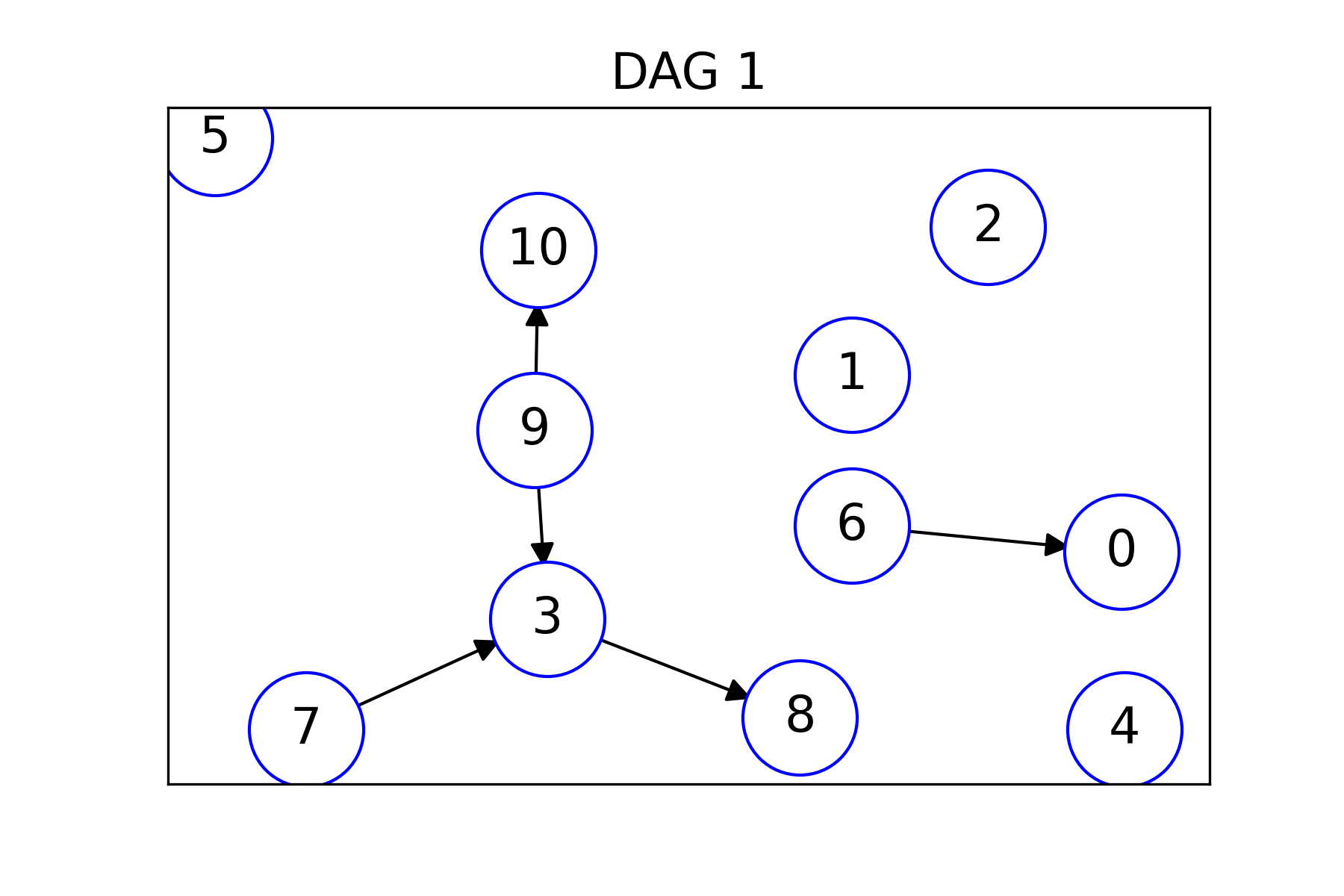}%
    \includegraphics[trim=10mm 10mm 10mm 5mm, clip, width=0.2\linewidth]{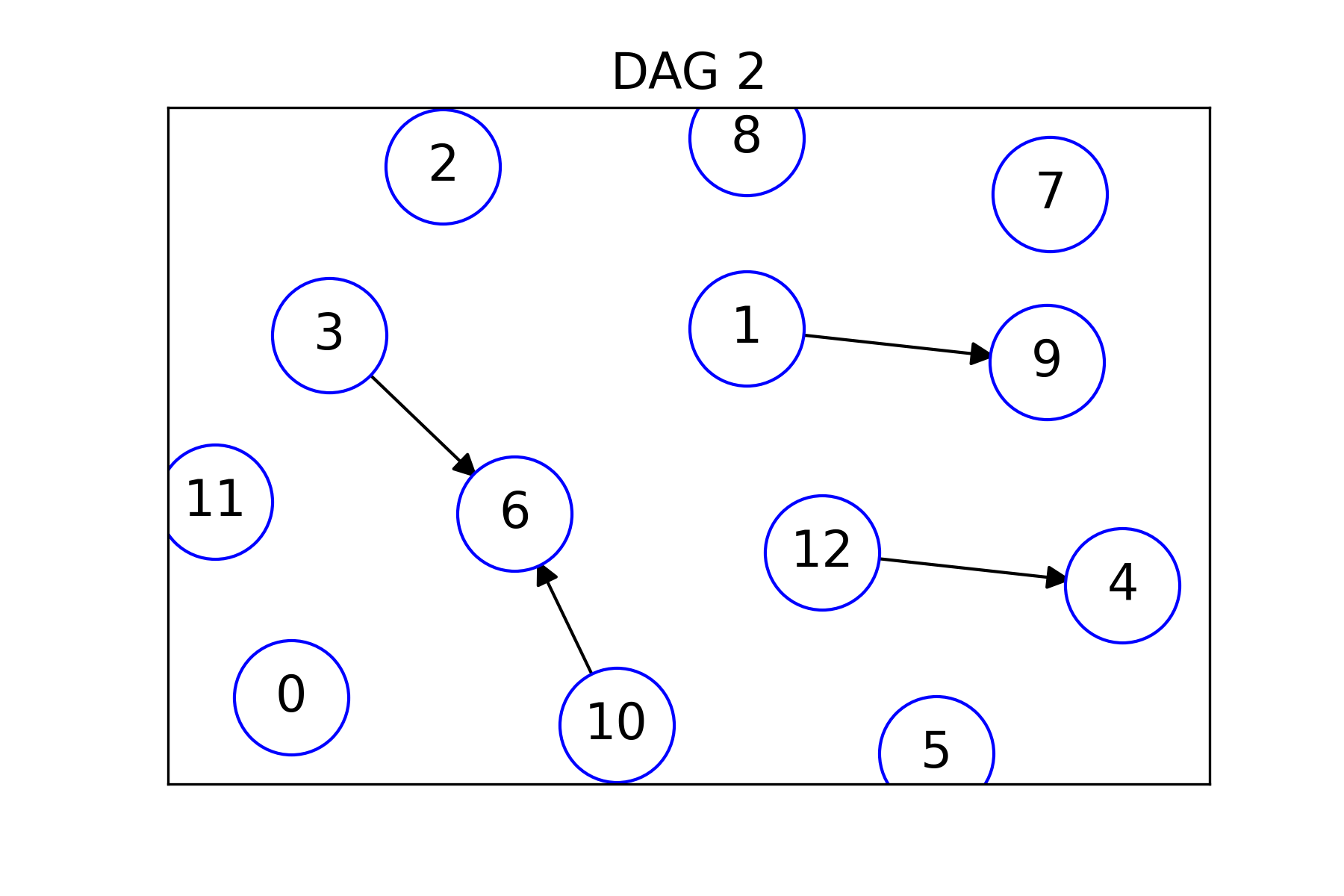}%
    \includegraphics[trim=10mm 10mm 10mm 5mm, clip, width=0.2\linewidth]{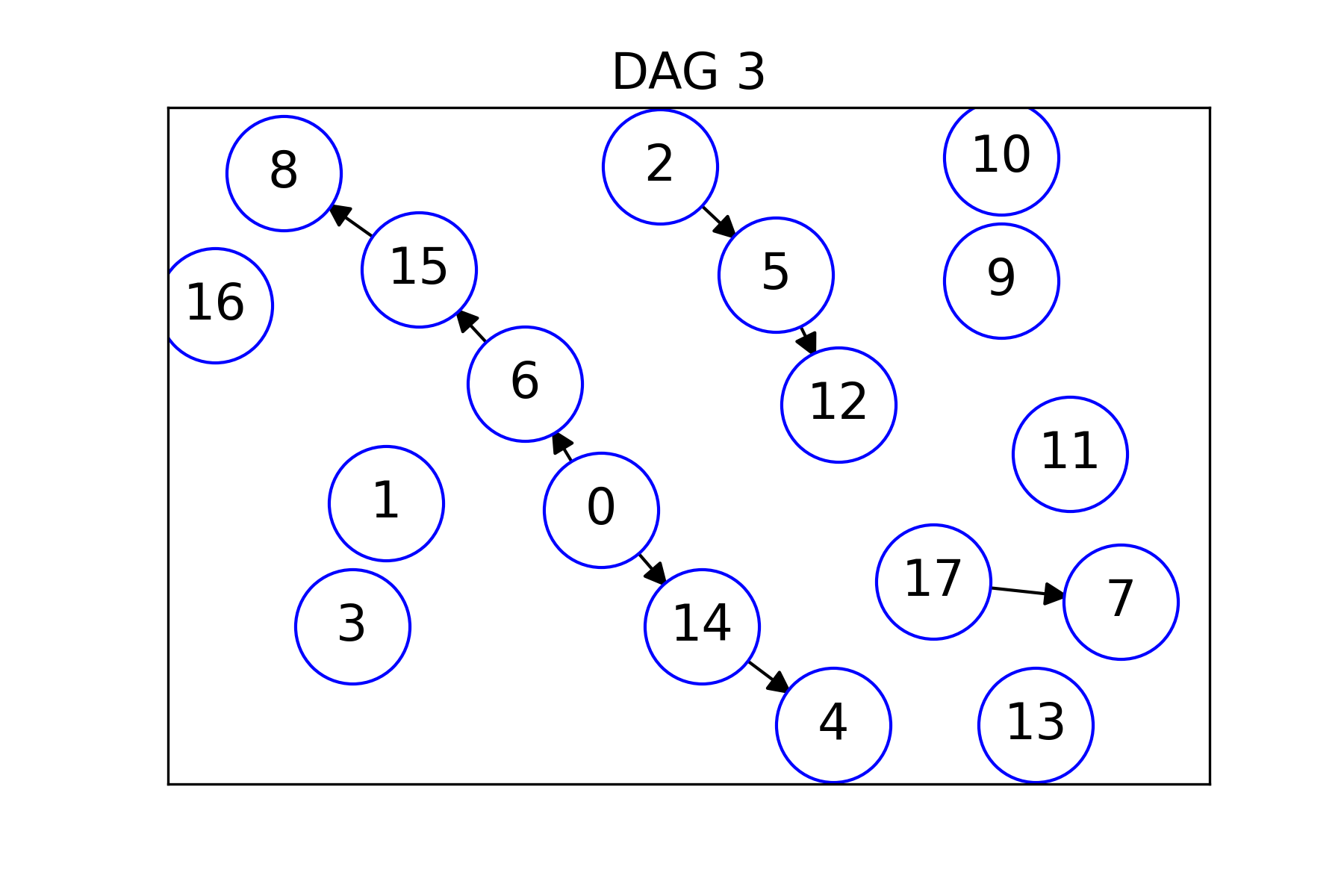}%
    \includegraphics[trim=10mm 10mm 10mm 5mm, clip, width=0.2\linewidth]{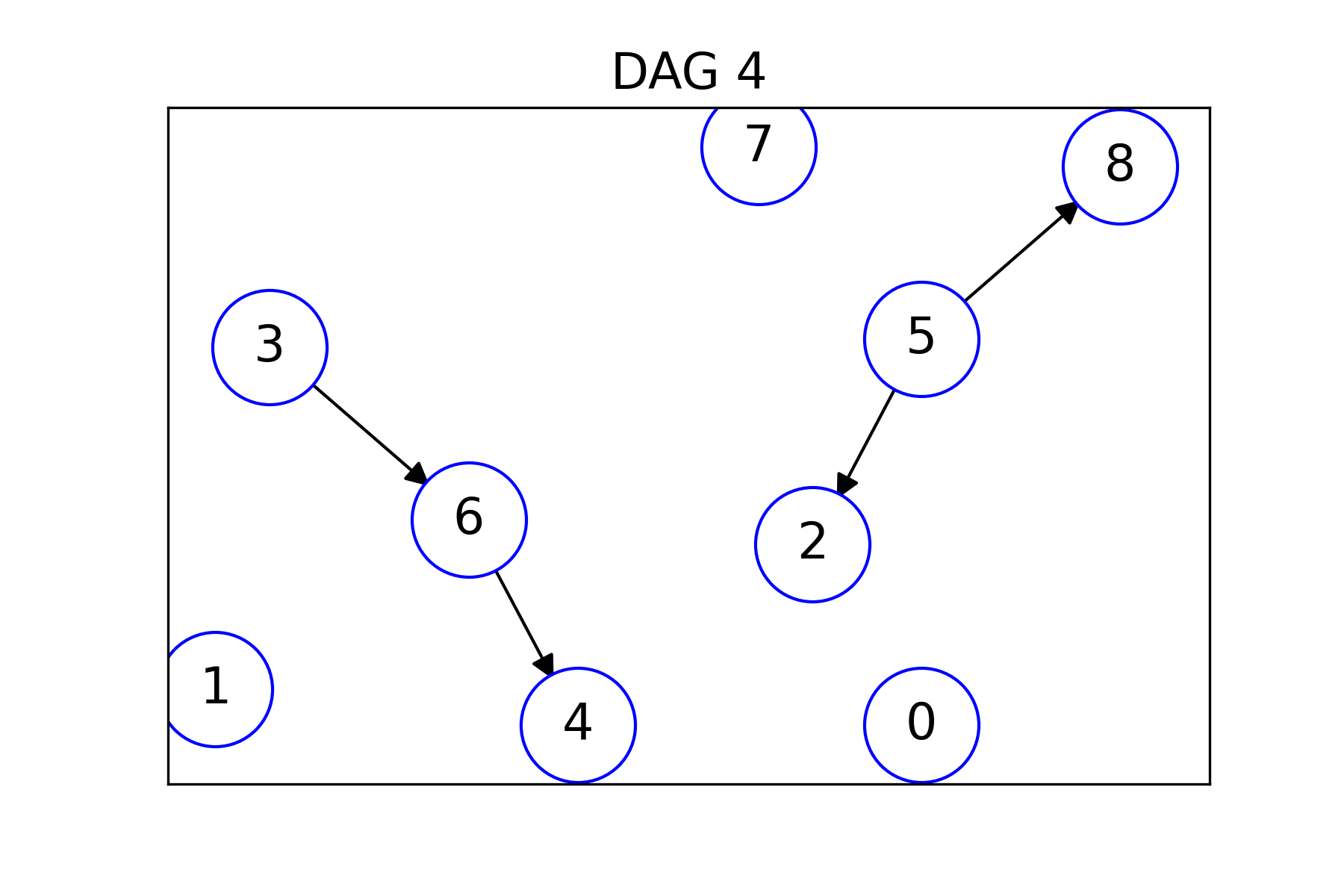}
    \includegraphics[trim=10mm 10mm 10mm 5mm, clip, width=0.2\linewidth]{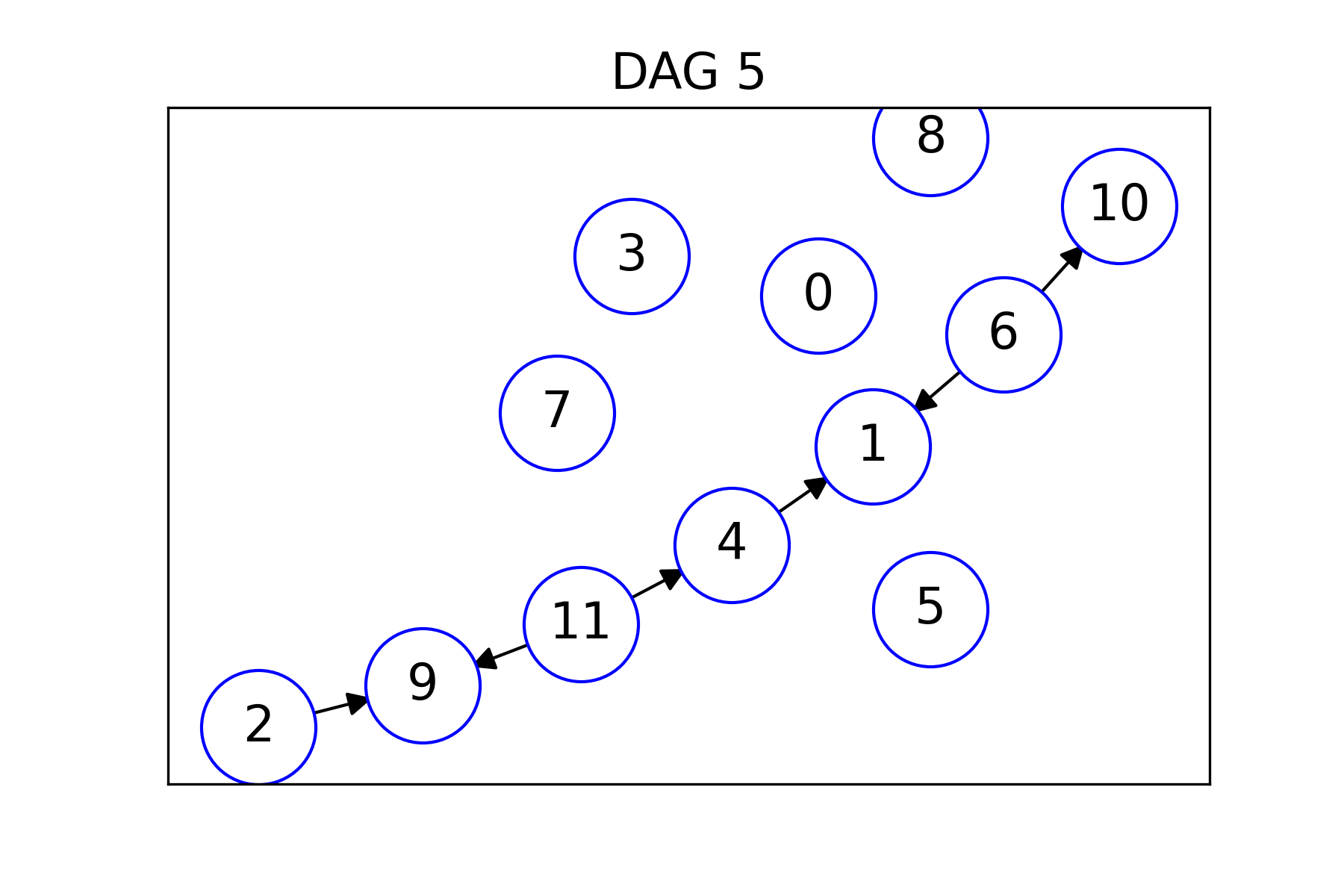}%
    \includegraphics[trim=10mm 10mm 10mm 5mm, clip, width=0.2\linewidth]{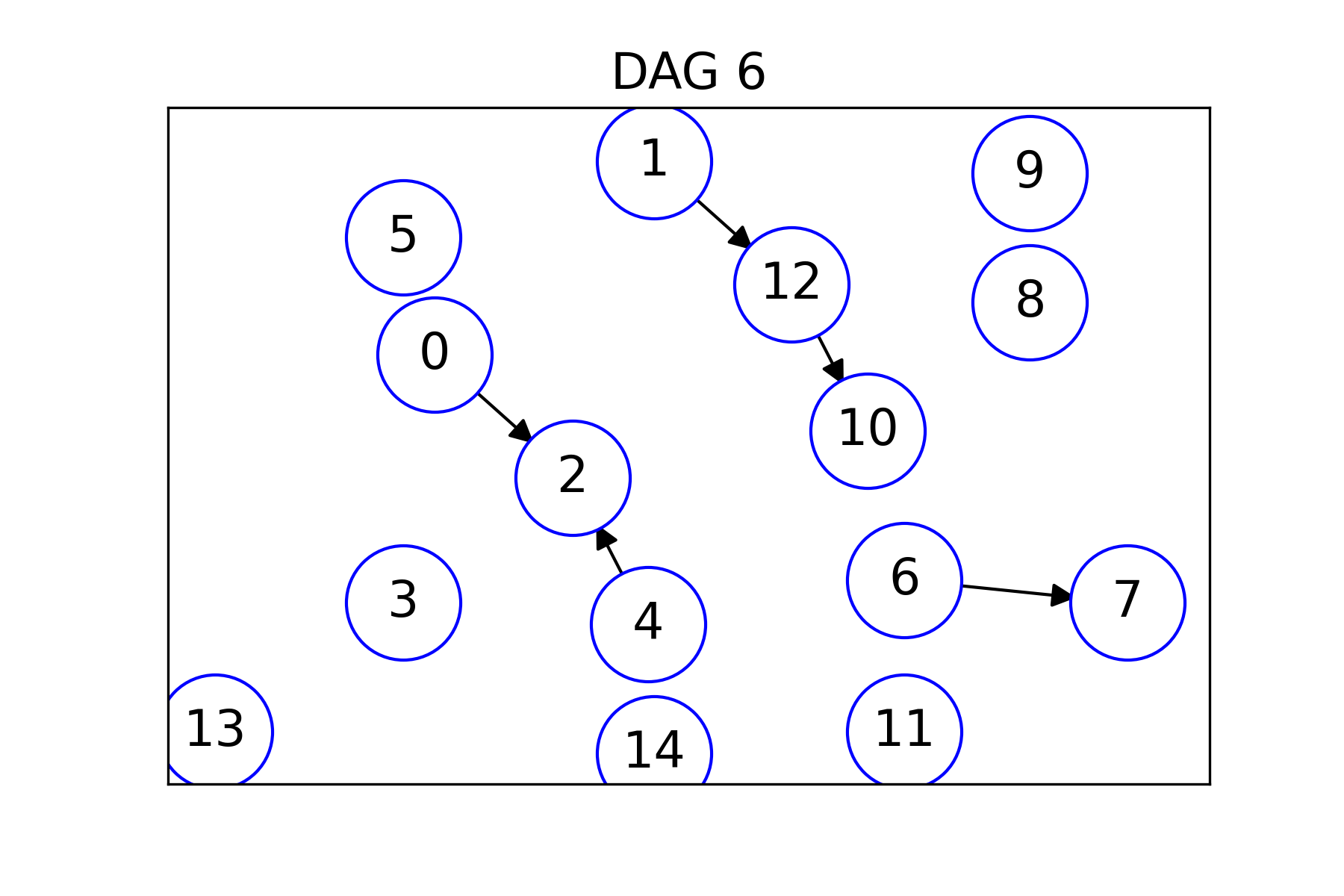}%
    \includegraphics[trim=10mm 10mm 10mm 5mm, clip, width=0.2\linewidth]{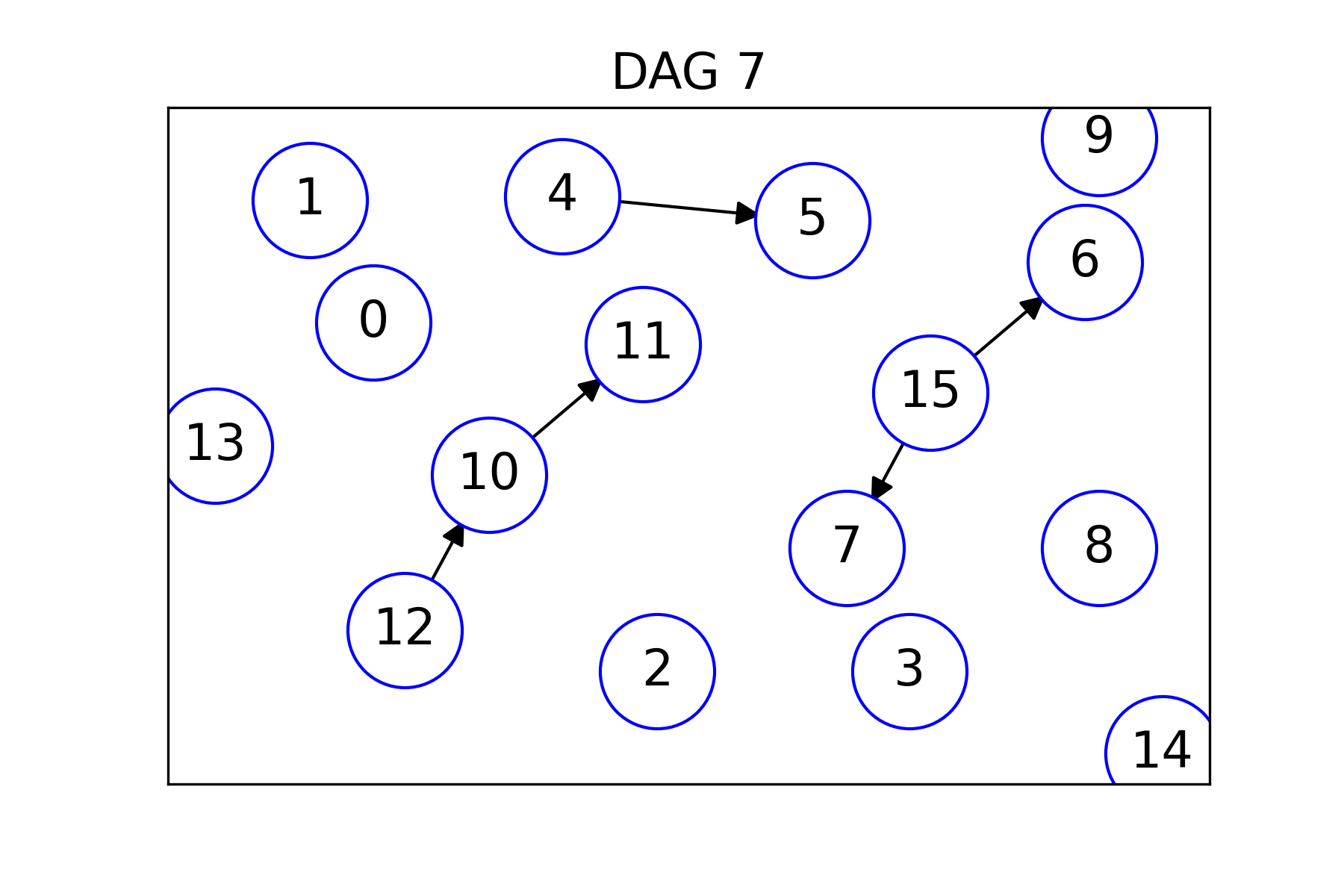}%
    \includegraphics[trim=10mm 10mm 10mm 5mm, clip, width=0.2\linewidth]{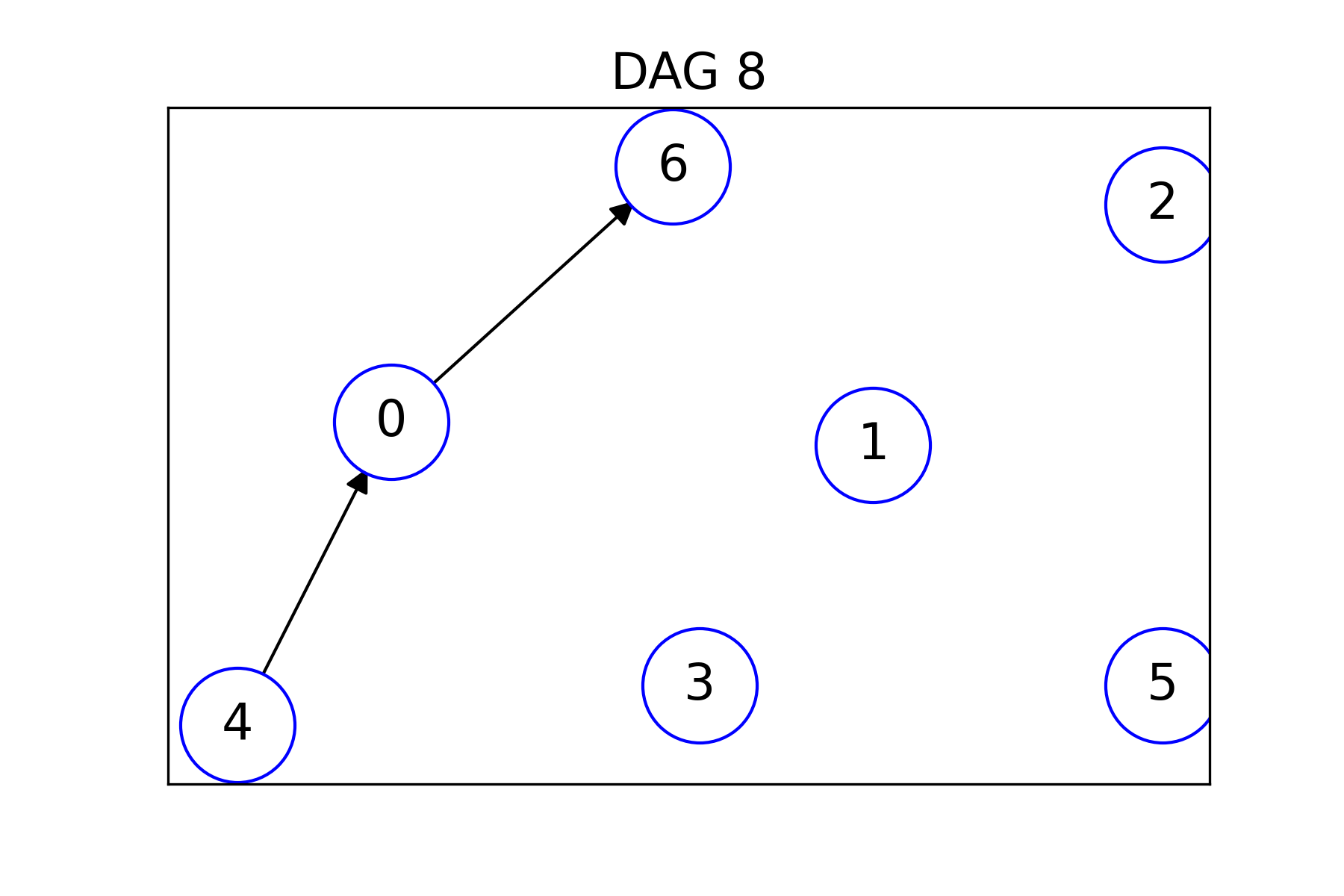}%
    \includegraphics[trim=10mm 10mm 10mm 5mm, clip, width=0.2\linewidth]{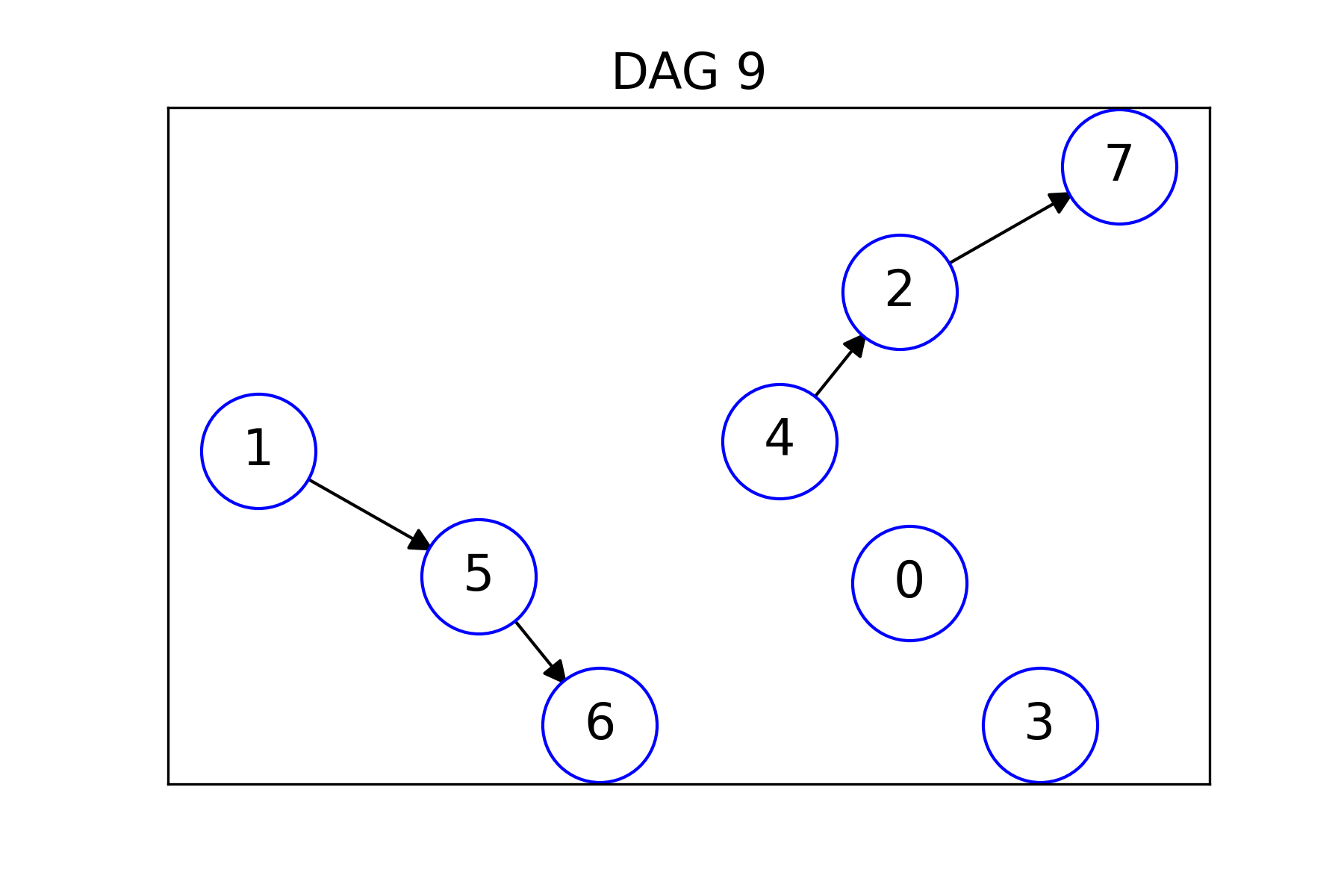}
    \includegraphics[trim=10mm 10mm 10mm 5mm, clip, width=0.2\linewidth]{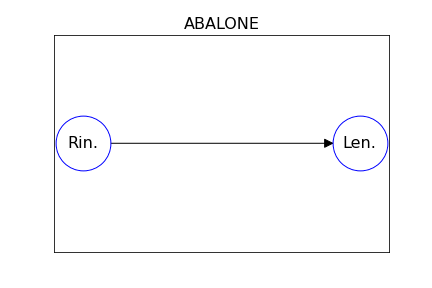}%
    \includegraphics[trim=10mm 10mm 10mm 5mm, clip, width=0.2\linewidth]{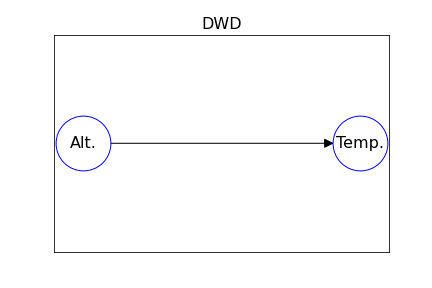}%
    \includegraphics[trim=10mm 10mm 10mm 5mm, clip, width=0.2\linewidth]{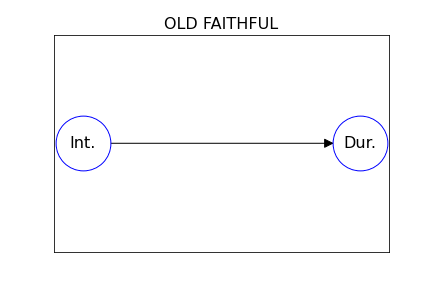}%
    \includegraphics[trim=10mm 10mm 10mm 5mm, clip, width=0.2\linewidth]{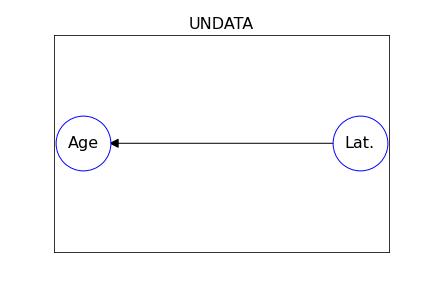}%
    \includegraphics[trim=10mm 10mm 10mm 5mm, clip, width=0.2\linewidth]{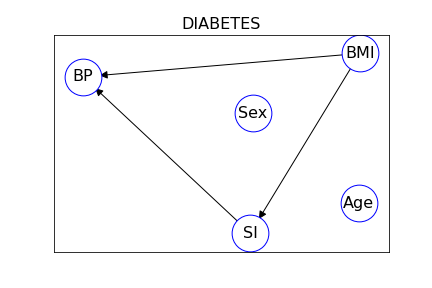}
    \caption{DAGs of synthetic datasets (first and second rows), and DAGs of real-world datasets (third row).}
    \label{fig:dag}
\end{figure*}

\subsection{Evaluation Measures}
\label{sec:eval_meas}
Since the contribution of this paper is twofold, i.e., the definition of an efficient and accurate method for nonlinear causal discovery and the design of a synthetic dataset generator, we need to evaluate and measure the validity of both aspects.

We establish to evaluate the correctness in the causal discovery task following the machine learning fashion~\cite{kalainathan2020causal}.
Let $G$ be the real DAG describing the causal structure, and $\tilde{G}$ the DAG inferred with a causal discovery approach.
We say that $G_{ij} = 1$ if in $G$ exists an edge from the node representing the feature $i$ to the node representing feature $j$, i.e., $i$ causes $j$ ($i \leftarrow j$).
Then, if $G_{ij} = \tilde{G}_{ij} = 1$ we have a True Positive, if $G_{ij} = \tilde{G}_{ij} = 0$ we have a True Negative, if $G_{ij} = 0 \wedge \tilde{G}_{ij} = 1$ a False Positive, and if $G_{ij} = 1 \wedge \tilde{G}_{ij} = 0$ a False Negative.
Given these definitions it is easy to define standard evaluation measures such as \textit{accuracy}, \textit{precision}, \textit{recall}, and f1~\cite{tan2005introduction}.

On the other hand, we evaluate the correctness of a synthetic dataset generative model using the following measures based on \textit{(i)} distances or \textit{(ii)} outlierness~\cite{guidotti2020data}. 
Let $X$ be the real dataset, and $\tilde{X}$ the synthetic one, we use the Sum of Squared Error (SSE) and Root Mean Squared Error (RMSE) as measures based on distances.
More in detail, for each feature $j$ we perform the Kernel Density Estimation\footnote{We used the \texttt{scikit-learn} KDE \url{https://scikit-learn.org/stable/modules/generated/sklearn.neighbors.KernelDensity.html}. In our experiments, we use the grid search for the bandwidth parameter in the interval $[-0.5, 1.5]$ with cross-validation with a Gaussian kernel. This allows us to choose the bandwidth whose score maximizes the log-likelihood of the KDE.} (KDE) on both $X^{(j)}$ and on $\tilde{X}^{(j)}$ to estimate the Probability Density Function (PDF).
We generate a set of 1000 random values according to the PDF, and we compare them using the SSE and the RMSE.
Finally, we aggregate the evaluations performed for the various features of the datasets by averaging them.
For estimating the number of outliers present in $\tilde{X}$ concerning $X$, we employ the Local Outlier Factor (LOF)~\cite{breunig2000lof}.
LOF is an outlier detection method that measures the local density deviation of a given instance and compares it to the local densities of its neighbors. 
Instances that have a density substantially lower than their neighbors are considered to be outliers.
In our experiments\footnote{We perform outlier detection with the LOF method as implemented by \texttt{sklearn} library: \url{https://scikit-learn.org/stable/modules/generated/sklearn.neighbors.LocalOutlierFactor.html}. We set the number of neighbors equal to 30 because it is typically set as a number higher than the minimum number of instances that a cluster must contain but lower than the maximum number of neighbor instances that can be potential outliers.} we check if any instance $\tilde{x}_i \in \tilde{X}$ can be considered an outlier concerning the real instances in $X$.
If the LOF is lower than one, it means that a higher density surrounds a point than its neighbors, and it is considered an inlier, i.e., an acceptable synthetic record in our setting.
On the other hand, the point is considered an outlier.

\subsection{Synthetic and Real Datasets}
\label{sec:synth_real_dataset}
In order to carefully perform the aforementioned evaluation, we require \textit{ground-truth} datasets of various dimensionalities with known causal relationships.
This aspect is fundamental in the context of causal discovery.
Indeed, to evaluate these methodologies, we need to rely on the structure of the DAG to test the identified causal relationships.
However, since the literature lacks this type of information, we developed a generator of random synthetic continuous datasets for which the causal structure is known a priori.

The generator first creates a random DAG $G$ to be used as ground truth.
The DAG $G$ is generated by selecting a number of random nodes in $[5, 20]$ and a number of random edges in $[2, \mathit{nbr\_nodes}/2]$.
Edges are assigned randomly to couples of nodes.
Then, it takes as input $G$ and returns a multivariate continuous dataset $X$ respecting the causal relationships where each column in $X$ represents a node in $G$.
The synthetic dataset $X$ is generated according to the following steps.
First, the features matching isolated and source nodes are generated, i.e., those modeling independent variables\footnote{
    Such distributions for independent variables are generated following one of these techniques.
    First approach: the distribution is a random uniform one with values in in $[\mathit{b}, \mathit{b}]$ with $b$ selected uniformly at random in $[5, 100]$.
    Second approach: the distribution is selected randomly among uniform, normal, exponential, log-normal, chisquare and beta.
    The parameters adopted are available on the repository.
    }.
Moreover, the generator adds a uniform noise in $[-1, 1]$ to each independent variable.
Following the topological ordering of the DAG, we ensure the independent variables are generated before the dependent ones.
Second, the features matching dependent variables are generated by combining the parent variables with randomly selected binary functions and by applying to each parent variable a randomly selected nonlinear function among sine, cosine, square root, logarithm, and tangent.
Finally, like for independent variables, the generator adds a uniform noise in $[-1, 1]$ also to dependent variables.

In our experiments, we generated 10 different DAGs illustrated in Figure~\ref{fig:dag} ($1^{\mathit{st}}, 2^{\mathit{nd}}$ rows).
For each DAG, we repeated the synthetic data generative procedure ten times, ending up with a total of 100 different synthetic datasets, each one with 1000 instances respecting the causal relationships.

We also experimented with real datasets typically used in papers of causal discovery for which the ground-truth DAG is known.
We selected \abalone{}, \old{} and \dwd{} from~\cite{hoyer2008nonlinear}, and \undata{} from~\cite{mooij2016distinguishing}.
In addition, since all these datasets are bivariate, we also considered the multivariate dataset \diabets{} for which we specified the ground-truth DAG.
The DAGs for these datasets are in Figure~\ref{fig:dag} ($3^{\mathit{rd}}$ row).

\begin{table}[t]
\centering
\caption{Runtime and F1-measure for causal discovery on DAGs with increasing number of features. $^*$ not all results considered.}
\label{tab:runtime_cd}
    \begin{tabular}{|c||cc|c||cc|}
    \hline
    \multirow{2}{*}{nbr features} & \multicolumn{3}{c||}{Time (sec)} & \multicolumn{2}{c|}{F1-measure} \\
     & \ncd{} & \ncda{} & {\sc apriori} & \ncd{} & \ncda{} \\
    \hline
    2 & $0.454$ & $0.462$ & $0.004$ & $1.000$ & $1.000$\\
    3 & $5.034$ & $0.464$ & $0.005$ & $1.000$ & $1.000$\\
    4 & $115.1$ & $0.990$ & $0.006$ & $0.890$ & $0.906$\\
    5 & $>3600$ & $1.033$ & $0.007$ & $-$ & $0.891$\\
    6 & $>3600$ & $1.229$ & $0.008$ & $-$ & $0.814$\\
    \hline
    avg & $>3600$ & $0.835$ & $0.006$ & $0.963^*$ & $0.926$\\
    \hline
    \end{tabular}
\end{table}

\begin{table}[t]
\centering
\caption{Runtime and F1-measure for causal discovery on real DAGs.}
\label{tab:runtime_cd_real}
    \begin{tabular}{|c||cc|c||cc|}
    \hline
    \multirow{2}{*}{nbr features} & \multicolumn{3}{c||}{Time (sec)} & \multicolumn{2}{c|}{F1-measure} \\
     & \ncd{} & \ncda{} & {\sc apriori} & \ncd{} & \ncda{} \\
    \hline
    \abalone{} & $0.208$ & $0.207$ & $0.008$ & $1.000$ & $1.000$\\
    \old{} & $0.089$ & $0.089$ & $0.006$ & $1.000$ & $1.000$\\
    \dwd{} & $0.058$ & $0.073$ & $0.008$ & $1.000$ & $1.000$\\
    \undata{} & $0.048$ & $0.045$ & $0.008$ & $1.000$ & $1.000$\\
    \diabets{} & $4607$ & $10.00$ & $0.009$ & $0.750$& $0.750$\\
    \hline
    \end{tabular}
\end{table}

\subsection{Baselines}
\label{sec:baselines}
We study the effectiveness of our proposal comparing it against some baselines and state-of-the-art proposals.

In particular, for the task of causal discovery, besides \ncd{}, we compare \ncda{} against coefficients typically used to detect correlations.
Indeed, our intuition is that correlation among variables is a clue for causation.
Therefore, simple coefficients like Pearson (\textsc{pc}), Spearman (\textsc{sc}) and Hoeffding's D (\textsc{hc})~\cite{tan2005introduction} could be used in replacement of Apriori.
We selected these three correlation indexes because they differ on \textit{(i)} the type of relationship which are able to recognize, \textit{(ii)} the direction of the relationship, i.e., monotonic vs. non-monotonic, \textit{(iii)} the statistic approach, i.e., parametric vs. non-parametric\footnote{We used the implementations of \url{https://docs.scipy.org/doc/scipy/reference/stats.html} and \url{https://github.com/PaulVanDev/HoeffdingD}.}.
For the evaluation of the Pearson and Spearman correlations, we checked the p-value using $0.05$ as a threshold, while for Hoeffding's D, we set the acceptance threshold to $0.03$.

For synthetic data generation, we compared against a random data generator (\rnd{}) that assumes uniform distribution and independence among all the variables.
Also, we compare \ncda{} against state-of-the-art data generators of Synthetic Data Vault library\footnote{\url{https://sdv.dev/SDV/index.html}. 
}.
We experimented with 
\tvae{}~\cite{xu2018synthesizing} and \ctgan{}~\cite{xu2019modeling} with default parameters generating 1000 instances.

\subsection{Experimental Settings}
\label{sec:exp_settings}
In the experiments we run \ncda{} and \gencda{} with the following parameters: $\mathit{n\_bins} \in [3, 10]$, $\mathit{min\_sup} \in [0.05, 0.4]$, $\mathit{max\_len} \in \{3, 4, 5\}$, $\alpha \in \{0.001, 0.01, 0.02, 0.05, 0.1\}$.
The default parameter justified by the experiments reported in Section~\ref{sec:exp_sensitivity_analysis} is $\mathit{n\_bins} = 10$, $\mathit{min\_sup}=0.05$, $\mathit{max\_len} = 3$, and $\alpha = 0.001$.
In \gencda{} as the list of distributions $D$ we consider the following among those available in \texttt{scipy}\footnote{\url{https://docs.scipy.org/doc/scipy/reference/stats.html}}: uniform, exponweib, expon, gamma, beta, alpha, chi, chi2, laplace, lognorm, norm, powerlaw.
A higher number of distributions (each one with its parameters) increases the computation time but also improves the performance as more accurate independent variables can be described.
For the ensemble regressor in \gencda{} we rely on the GPR, SVM, kNN, and DTR of \texttt{scikit-learn} trained with default parameters. 

\begin{table}[t]
\centering
\caption{Comparison of \ncda{} with correlation coefficients to detect causalities on synthetic DAGs. Best results in bold.}
\label{tab:perf_cd}
\setlength{\tabcolsep}{1mm}
    \begin{tabular}{|c|cccc|cccc|cccc|}
    \hline
    \multirow{2}{*}{DAG} & \multicolumn{4}{c|}{Accuracy} & \multicolumn{4}{c|}{Precision} & \multicolumn{4}{c|}{Recall} \\
     & {\sc pc} & {\sc sc} & {\sc hc} & \ncda{} & {\sc pc} & {\sc sc} & {\sc hc} & \ncda{} & {\sc pc} & {\sc sc} & {\sc hc} & \ncda{}\\
     \hline
    0 & .89 & .88 & \textbf{.93} & .91 & .67 & .60 & .86 &   \textbf{.88} & .62 & \textbf{.68} & .60 & .50 \\
    1  & .87 & .87 & .92 & \textbf{.94} &  .32 &  .31 &  .66 & \textbf{.88} &  .38 &  .40 &  .42 &  \textbf{.44} \\
    2 &  .92 &  .91 &  .95 &  \textbf{.97} & .23 & .21 & .63 & \textbf{1.0} &  .23 &  .25 &  .28 &  \textbf{.42} \\
    3 &  .88 &  .87 &  .92 &  \textbf{.97} & .10 & .10 & .14 & \textbf{.92} &  .17 &  .19 &  .12 &  \textbf{.53} \\
    4 &  .93 &  .90 &  \textbf{.95} &  \textbf{.95} & .68 & .54 & .83 & .\textbf{89} &  .70 &  \textbf{.88} &  .78 &  .70 \\
    5 &  .88 &  .87 &  .92 &  \textbf{.93} & .32 & .30 &   .64 & \textbf{.88} &  .23 &  \textbf{.25} &  \textbf{.25} &  \textbf{.25} \\
    6 &  .90 &  .90 &  .90 &  \textbf{.97} & .16 & .15 &   .38 & \textbf{.93} &  .20 &  .28 &  .29 &  .52 \\
    7 &  .90 &  .89 &  .94 &  \textbf{.98} & .03 & .04 &   .07 & \textbf{.94} &  .04 &  .08 &  .04 &  .\textbf{50} \\
    8 &  .92 &  .93 &  \textbf{.97} &  .96 & .61 & .60 &   .80 & \textbf{.87} &  .95 &  \textbf{1.0} &  .95 &  .80 \\
    9  &  .88 &  .89 &  \textbf{.93} &  .92 & .63 & .63 & .89 & \textbf{.97} &  .57 &  \textbf{.60} &  \textbf{.60} &  .45 \\
    \hline
    avg & .89 & .89 & .93 & \textbf{.95} & .37 & .34 & .58 & \textbf{.91} & .41 & .45 & .43 & \textbf{.51} \\
    \hline
    \end{tabular}
\end{table}

\begin{figure}[t]
\centering
    \includegraphics[trim=25mm 5mm 25mm 2mm, clip,width=.5\linewidth]{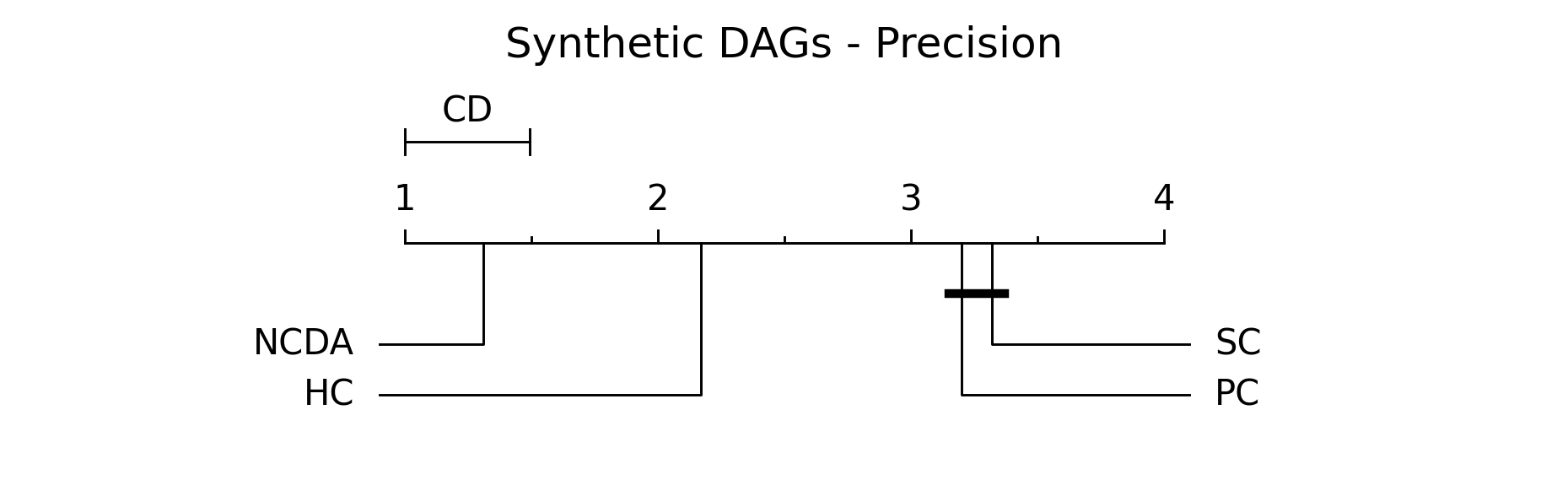}%
    \includegraphics[trim=25mm 5mm 25mm 2mm, clip,width=.5\linewidth]{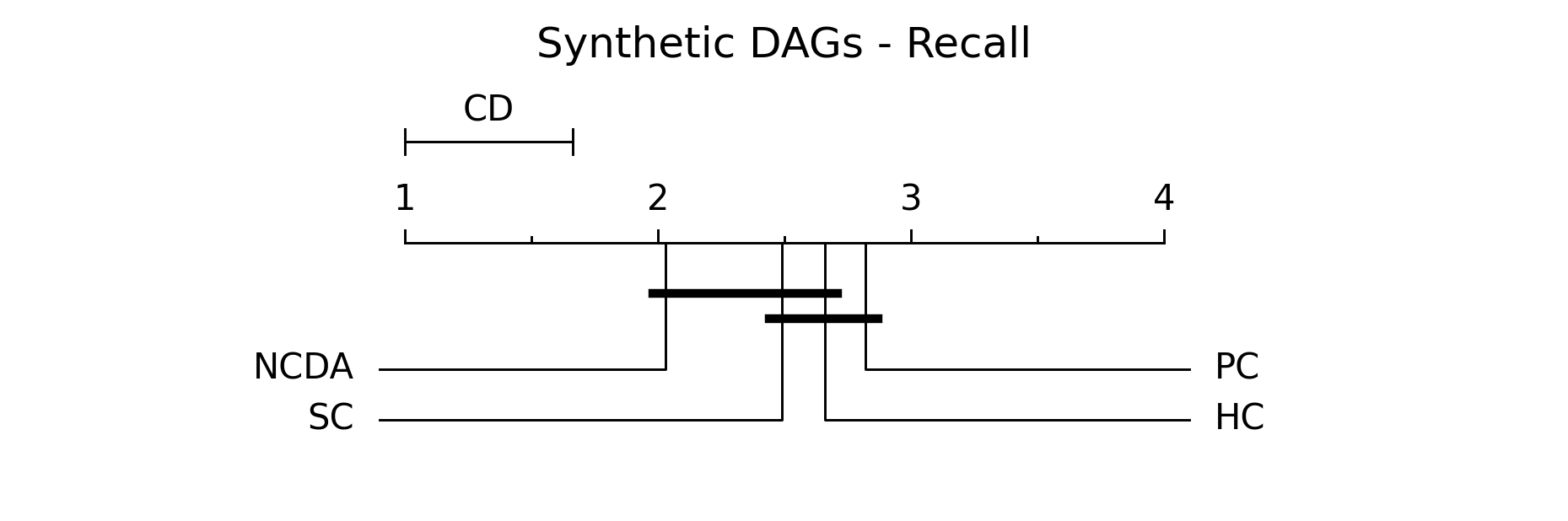} 
    \caption{CD plots with Nemenyi at 95\% confidence level 
    or synthetic DAGs 
    }
    \label{fig:cdplot_perf_cd}
\end{figure}

\begin{table}[t]
\centering
\caption{Comparison of \ncda{} with correlation coefficients to detect causalities on real DAGs.}
\label{tab:perf_cd_real}
\setlength{\tabcolsep}{0.8mm}
    \begin{tabular}{|c|cccc|cccc|cccc|}
    \hline
    \multirow{2}{*}{DAG} & \multicolumn{4}{c|}{Accuracy} & \multicolumn{4}{c|}{Precision} & \multicolumn{4}{c|}{Recall} \\
     & {\sc pc} & {\sc sc} & {\sc hc} & \ncda{} & {\sc pc} & {\sc sc} & {\sc hc} & \ncda{} & {\sc pc} & {\sc sc} & {\sc hc} & \ncda{}\\
     \hline
    \abalone{} & $1.0$ & $1.0$ & $1.0$ & $1.0$ & $1.0$ & $1.0$ & $1.0$ & $1.0$ & $1.0$ & $1.0$ & $1.0$ & $1.0$ \\
    \old{}  & $1.0$ & $1.0$ & $1.0$ & $1.0$ & $1.0$ & $1.0$ & $1.0$ & $1.0$ & $1.0$ & $1.0$ & $1.0$ & $1.0$ \\
    \dwd{} & $1.0$ & $1.0$ & $1.0$ & $1.0$ & $1.0$ & $1.0$ & $1.0$ & $1.0$ & $1.0$ & $1.0$ & $1.0$ & $1.0$ \\
    \undata{} & $1.0$ & $1.0$ & $1.0$ & $1.0$ & $1.0$ & $1.0$ & $1.0$ & $1.0$ & $1.0$ & $1.0$ & $1.0$ & $1.0$ \\
    \diabets{} & $0.4$ & $0.4$ & $0.8$ & $0.9$ & $0.3$ & $0.3$ & $0.6$ & $0.6$ & $1.0$ & $1.0$ & $0.6$ & $1.0$ \\
    \hline
    \end{tabular}
\end{table}

\subsection{Results}
\label{sec:exp_results}
The first aspect that we analyze is the impact of \textsc{apriori} on the performance of \ncd{} for the task of causal discovery.
In Table~\ref{tab:runtime_cd} we observe the performance of \ncd{} and \ncda{} in terms of runtime and F1-measure for DAGs with a growing number of features.
With $^*$ we indicate that for \ncd{} the average value considers only the cases in which the procedure terminated within an hour.
Given a \textit{number of features}, we randomly generated 50 DAGs and datasets with the approach of Sec.~\ref{sec:synth_real_dataset}.
We notice how \ncda{} has a remarkable improvement in terms of runtime that becomes evident when \textit{number of features} is higher than 4.
Thus, the computational time required by \ncda{} is exponentially lower than the one required by \ncd{}.
The third column shows the negligible impact of \textsc{apriori} on the runtime of \ncda{}.
Besides, from the F1-measure, we notice that \textsc{apriori} does not impact the performance of \ncda{} to \ncd{} when observing the correctness of the causal relationships discovered in terms of F1-measure.
Similar results are on Table~\ref{tab:runtime_cd_real} for the real datasets.

In Table~\ref{tab:perf_cd} we report the accuracy, precision and recall for the task of causal discovery for the ten synthetic DAGS of Figure~\ref{fig:dag} ($1^{st}$ and $2^{nd}$ rows) comparing \ncda{} against the correlation indexes Pearson (\textsc{pc}), Spearman (\textsc{sc}) and Hoeffding's D (\textsc{hc}).
We remark that for each DAG, we generate ten different datasets.
In Table~\ref{tab:perf_cd} we report the average performance among the ten datasets.
We immediately notice that \ncda{} has the overall better accuracy.
However, accuracy cannot be very informative due to the high number of true negative related to the sparseness of the DAGs.
Concerning precision is the best performer on all the DAGs.
This aspect is crucial as the causalities identified result to be correct nearly always.
Finally, recall \ncda{} is more conservative as it sometimes fails to recognize some causal relationships to correlation indexes.
However, the average recall remains higher than the competitors.

\begin{table}[t]
\centering
\caption{Comparison of \gencda{} with generative approaches on synthetic datasets: error measures. Best results in bold.}
\label{tab:perf_dg_dist}
\setlength{\tabcolsep}{1.2mm}
    \begin{tabular}{|c|cccc|cccc|}
    \hline
    \multirow{2}{*}{DAG} & \multicolumn{4}{c|}{SSE} & \multicolumn{4}{c|}{RMSE} \\
     & \rnd{} & \tvae{} & \ctgan{} & \gencda{} & \rnd{} & \tvae{} & \ctgan{} & \gencda{} \\
     \hline
    0 & $.629$ & \textbf{.137} & $.321$ & $.257$ & $.018$ & $.024$ & $.036$ & \textbf{.012} \\
    1 & $.386$ & $.156$ & \textbf{.136} & $.138$ & $.012$ & $.022$ & $.021$ & \textbf{.006} \\
    2 & $.320$ & $.176$ & $.151$ & \textbf{.082} & $.011$ & $.025$ & $.023$ & \textbf{.006} \\
    3 & $.496$ & \textbf{.178} & $.191$ & $.205$ & $.016$ & $.028$ & $.029$ & \textbf{.009} \\
    4 & $.525$ & \textbf{.091} & $.147$ & $.172$ & $.016$ & $.020$ & $.025$ & \textbf{.008} \\
    5 & $.248$ & $.095$ & $.102$ & \textbf{.074} & $.010$ & $.019$ & $.017$ & \textbf{.005} \\
    6 & $.431$ & \textbf{.114} & $.139$ & $.167$ & $.013$ & $.021$ & $.021$ & \textbf{.007} \\
    7 & $.376$ & $.197$ & $.122$ & \textbf{.137} & $.013$ & $.026$ & $.021$ & \textbf{.007} \\
    8 & $.411$ & \textbf{.229} & $.206$ & $.283$ & $.014$ & $.029$ & $.029$ & \textbf{.009} \\
    9 & $.435$ & $.108$ & $.184$ & \textbf{.084} & $.015$ & $.023$ & $.026$ & \textbf{.006} \\
    \hline
    avg & $.426$ & \textbf{.148} & $.169$ & $.160$ & $.014$ & $.024$ & $.025$ & \textbf{.007} \\
    \hline
    \end{tabular}
\end{table}

\begin{table}[t]
\centering
\caption{Comparison of \gencda{} with generative approaches on synthetic datasets: outlier measures. Best results in bold.}
\label{tab:perf_dg_out}
\setlength{\tabcolsep}{1mm}
    \begin{tabular}{|c|cccc|cccc|}
    \hline
    \multirow{2}{*}{DAG} & \multicolumn{4}{c|}{LOF} & \multicolumn{4}{c|}{\# Outliers} \\
     & \rnd{} & \tvae{} & \ctgan{} & \gencda{} & \rnd{} & \tvae{} & \ctgan{} & \gencda{} \\
     \hline
    0 & $0.660$ & $0.459$ & $0.673$ & \textbf{0.423} & $387$ & \textbf{3} & $295$ & $44$ \\
    1 & $324.4$ & \textbf{0.457} & $10.59$ & $0.463$ & $915$ & \textbf{131} & $691$ & $292$ \\
    2 & $38.03$ & $0.448$ & \textbf{0.322} & $0.443$ & $982$ & \textbf{107} & $424$ & $151$ \\
    3 & $>100$ & \textbf{0.471} & $>100$ & $0.480$ & $888$ & \textbf{125} & $569$ & $232$ \\
    4 & $3.078$ & $0.462$ & $1.323$ & \textbf{0.440} & $658$ & \textbf{12} & $561$ & $156$ \\
    5 & $>100$ & $>100$ & $>100$ & \textbf{0.370} & $982$ & $454$ & $704$ & \textbf{405} \\
    6 & $22.67$ & $0.450$ & \textbf{0.377} & $0.480$ & $845$ & \textbf{14}  & $428$ & $181$ \\
    7 & $>100$ & $>100$ & $>100$ & $0.480$ & $903$ & $407$ & $533$ & \textbf{123} \\
    8 & $0.476$ & \textbf{0.464} & $0.437$ & $0.470$ & $244$ & \textbf{2} & $270$ & $86$ \\
    9 & $2.667$ & $0.463$ & $2.008$ & \textbf{0.442} & $558$ & \textbf{56} & $547$ & $116$ \\
    \hline
    med & $7.177^*$ & $0.460^*$ & $0.660^*$ & \textbf{0.461} & $736$ & \textbf{27} & $502$ & $179$ \\
    \hline
    \end{tabular}
\end{table}

We analyze Table~\ref{tab:perf_cd} with the non-parametric Friedman test that compares the average ranks of the causal discovery methods over multiple datasets w.r.t. the various evaluation measure. 
The null hypothesis that all methods are equivalent is rejected with $\mathit{p{-}value} {<} 0.0001$ for all the measures observed.
The comparison of the ranks of all methods against each other is visually represented in Figure~\ref{fig:cdplot_perf_cd} with Critical Difference (CD) diagrams~\cite{demsar2006statistical}
Two methods are tied if the null hypothesis that their performance is the same cannot be rejected using the Nemenyi test at $\alpha{=}0.05$. 
\ncda{} has the best rank for precision with statistically significant performance.
On the other hand, ranks are not statistically significant w.r.t. recall, but \ncda{} remains the best performer for these metrics.

In Table~\ref{tab:perf_cd_real} we report accuracy, precision and recall for the causal discovery task on the five real DAGs of Figure~\ref{fig:dag} ($3^{rd}$ row) comparing \ncda{} against the correlation indexes Pearson (\textsc{pc}), Spearman (\textsc{sc}) and Hoeffding's D (\textsc{hc}).
Analyzing the results we notice that for the bivariate datasets \abalone{}, \old{}, \dwd{} and \undata{}, all the approaches manage to identify the causal direction. 
However, these good results do not indicate that it is possible to identify causalities by exploiting correlations since this only happens when there is only a single causal dependence. 
Indeed, for the \diabets{} dataset Pearson and Spearman have bad performance.
Hoeffding has good precision and a good recall, but none of them is perfect.
Finally, \ncda{} obtains the same perfect results on bivariate datasets and overcomes Hoeffding on \diabets{} as Hoeffding has an F1 of $0.6$ while \ncda{} of $0.75$.
The non-parametric Friedman test confirms the statistical significance of the results with a $\mathit{p{-}value} {<} 0.0005$ for all the measures observed.

In Tables~\ref{tab:perf_dg_dist}, \ref{tab:perf_dg_out} and Tables~\ref{tab:perf_dg_dist_real}, \ref{tab:perf_dg_outliers_real} we report the evaluation for the data generation task for synthetic and real datasets, respectively.
We highlight that for all these Tables, the non-parametric Friedman test confirms the statistical significance of the results with a $\mathit{p{-}value} {<} 0.0001$ for all the measures observed.
Besides the measures reported in these Tables and discussed in the following, it is worth mentioning that the runtimes of the generation methods are comparable on the relatively small datasets analyzed,
Indeed, the average runtime in seconds for generating synthetic data is $10.3$ for \rnd{}, $19.6$ for \tvae{}, $22.7$ for \ctgan{}, and $16.7$ for \gencda{}.
We notice that, as expected, \rnd{} is the fastest approach while \tvae{} and \ctgan{} are the slowest.
\gencda{} is the second-fastest performer, which is a valuable property considering the good qualitative results discussed in the following.

\begin{figure}[t]
\centering
    \includegraphics[trim=20mm 5mm 20mm 2mm, clip, width=.5\linewidth]{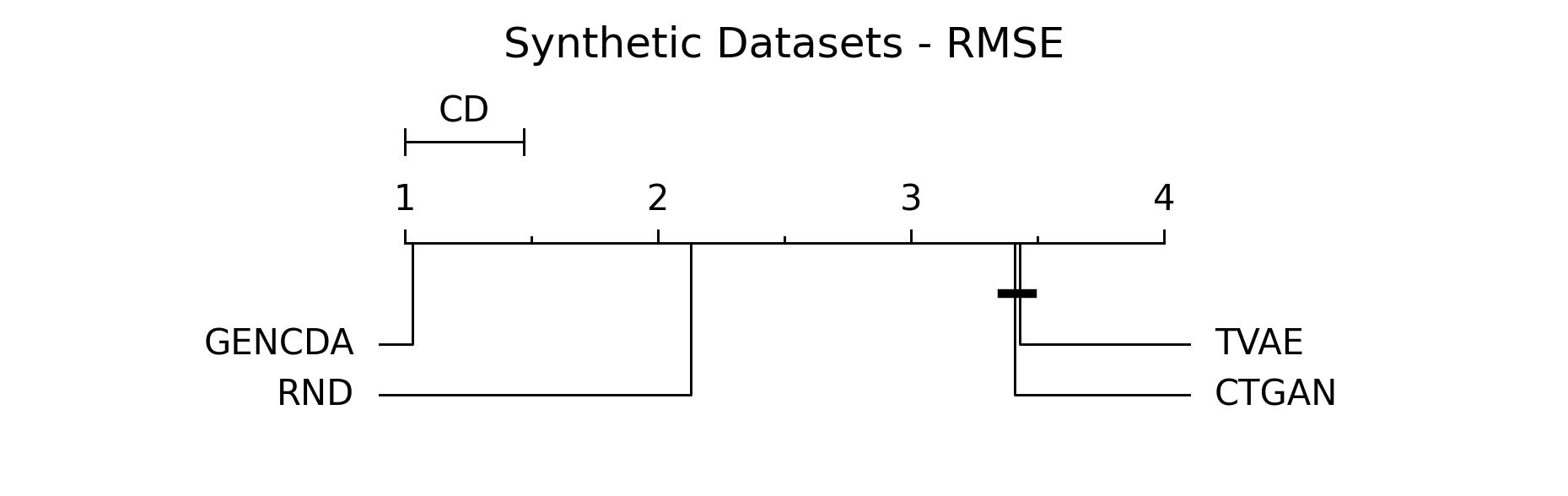}%
    \includegraphics[trim=20mm 5mm 20mm 2mm, clip, width=.5\linewidth]{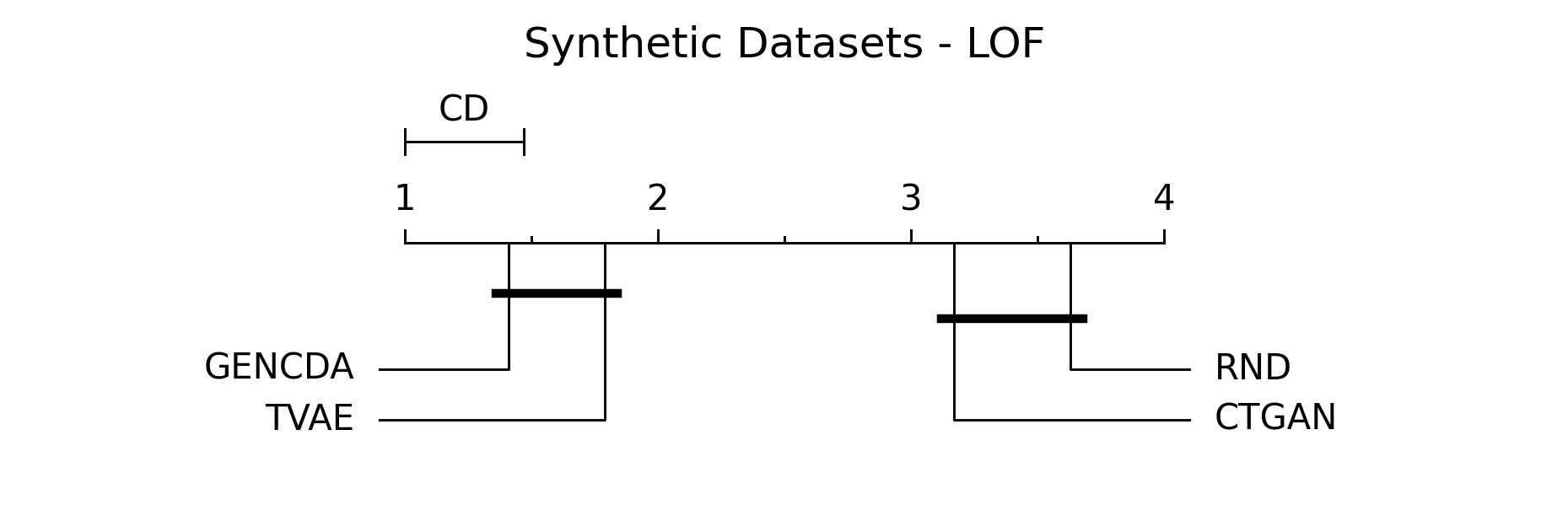} 
    \includegraphics[trim=20mm 5mm 20mm 2mm, clip, width=.5\linewidth]{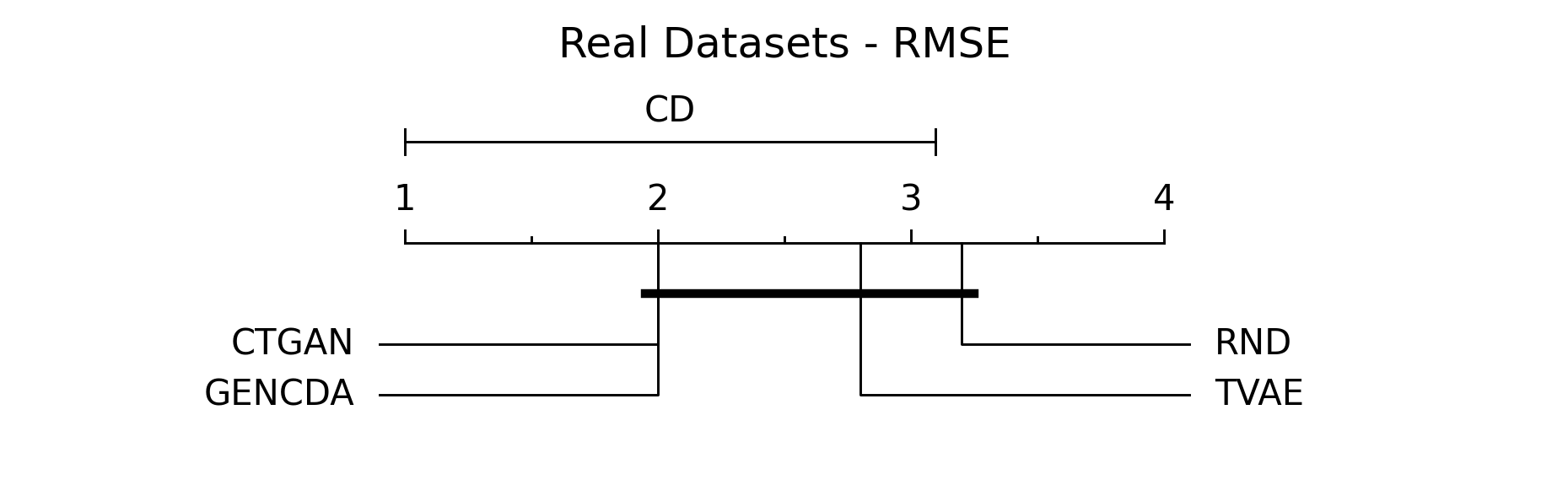}%
    \includegraphics[trim=20mm 5mm 20mm 2mm, clip, width=.5\linewidth]{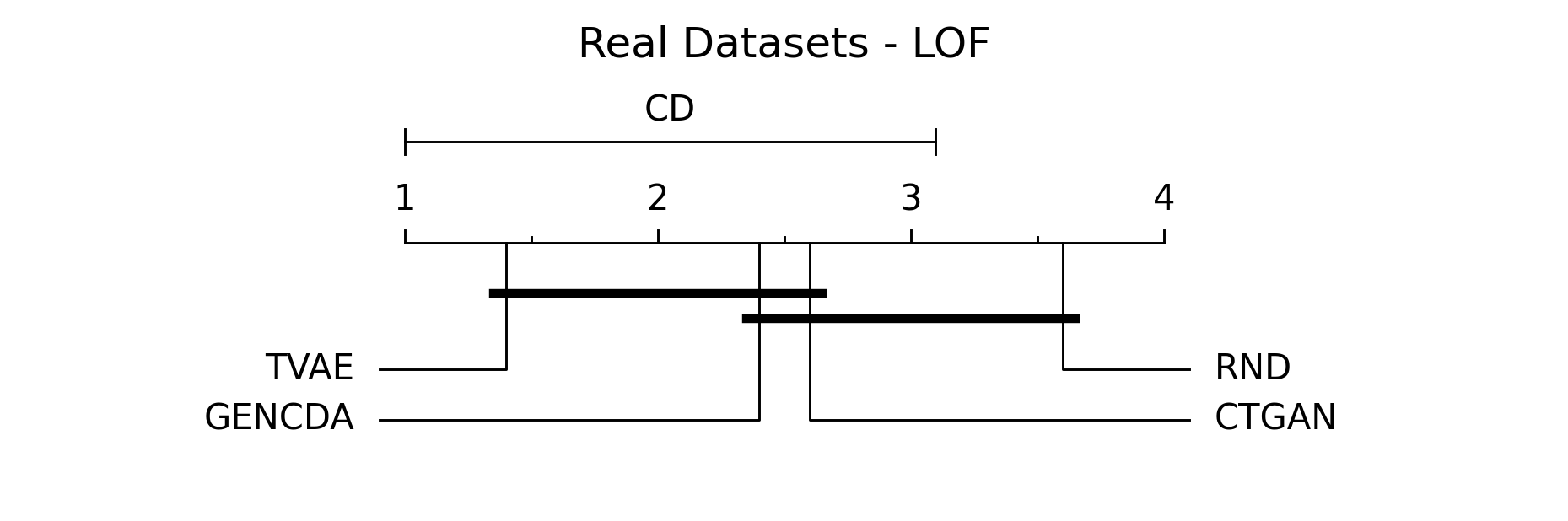} 
    \caption{CD plots with Nemenyi at 95\% confidence level. 
    }
    \label{fig:cdplot_perf_dg}
\end{figure}

In Tables~\ref{tab:perf_dg_dist} and~\ref{tab:perf_dg_dist_real} we observe the performance in terms of error measures (SSE and RMSE, the lower the better) for synthetic and the real datasets obtained by comparing the data distributions as detailed in Section~\ref{sec:eval_meas}.
Table~\ref{tab:perf_dg_dist} shows the mean values obtained for the different runs among the ten datasets generated for each DAG.
We notice that \gencda{} is the best performer in terms of RMSE for synthetic datasets and the second-best performer in terms of SSE.
Indeed, \tvae{} has very good results followed by \ctgan{} and finally by \rnd{}.
Therefore, it seems that the neural networks modeling the VAE learned by \tvae{} are somewhat able to capture also the causalities learned by \gencda{} through the \ncda{} procedure.
However, concerning \tvae{}, \gencda{} shows a smaller running time and therefore higher usability on a larger dataset.
This is due to the fact that \gencda{} learns relationships among variables exploiting patterns and does not need to consider many instances as required by VAEs in \tvae{}.
The CD plots on the left in Figure~\ref{fig:cdplot_perf_dg} validate these observations: \gencda{} is statistically the best performer for the synthetic datasets while it is comparable with all the others.

In Tables~\ref{tab:perf_dg_out} and~\ref{tab:perf_dg_outliers_real} we report the performance in terms of LOF score and of the number of outliers (the lower, the better).
We calculated the number of outliers as the number of synthetically generated instances for which LOF is higher than one\footnote{This choice is driven by the library used to calculate this measure.}.
Table~\ref{tab:perf_dg_out} shows the median values for the different runs among the ten datasets generated for each DAG.
For LOF, we write $>100$ when the median score is very big.
This indicates that more than half of the records synthetically generated are considered outliers by LOF\footnote{These results also depends on the parameter setting of LOF.}.
Thus, the total median values in the last line of Table~\ref{tab:perf_dg_out} have an asterisk $(*)$ when the aggregation is done without considering these very high values.
We immediately notice that \gencda{} is the only method without asterisks indicating that the majority of the population generated has a low LOF: the synthetic records of \gencda{} are fewer outliers than those generated with other methods.
Again, these results are underlined by the CD plots on the top right in Figure~\ref{fig:cdplot_perf_dg}: \gencda{} is the best performer followed by \tvae{} and they are statistically comparable.
However, empirically \gencda{} shows better results for synthetic datasets.
On the other hand, for real datasets, the performance is comparable among the various methods, but \gencda{} is still ranked first.
Concerning the number of outliers, results are comparable but, with the above parameter setting of LOF, \tvae{} generates a slightly lower number of outliers than \gencda{} for synthetic datasets, while the results are comparable for real ones.

\begin{table}[t]
\centering
\caption{Comparison of \gencda{} with generative approaches on real datasets: error-based measures.}
\label{tab:perf_dg_dist_real}
\setlength{\tabcolsep}{1mm}
    \begin{tabular}{|c|cccc|cccc|}
    \hline
    \multirow{2}{*}{DAG} & \multicolumn{4}{c|}{SSE} & \multicolumn{4}{c|}{RMSE} \\
     & \rnd{} & \tvae{} & \ctgan{} & \gencda{} & \rnd{} & \tvae{} & \ctgan{} & \gencda{} \\
     \hline
    \abalone{} & $.471$ & $.080$ & $.067$ & \textbf{.035} & $.016$ & $.025$ & $.023$ & \textbf{.005} \\
    \old{} & $6.05$ & \textbf{.066} & $.116$ & $5.04$ & $.059$ & $.020$ & $.026$ & \textbf{.056} \\
    \dwd{} & $.469$ & \textbf{.075} & $.462$ & $.185$ & $.007$ & $.019$ & $.015$ & \textbf{.009} \\
    \undata{} & $.125$ & \textbf{.019} & $.006$ & $.051$ & $.010$ & $.013$ & $.006$ & \textbf{.006} \\
    \diabets{} & $226$ & \textbf{.000} & $.004$ & $213$ & $1.48$ & \textbf{.001} & $.005$ & $1.15$ \\
    \hline
    \end{tabular}
\end{table}

\begin{table}[t]
\centering
\caption{Comparison of \gencda{} with generative approaches on real datasets: outlier-based measures.}
\label{tab:perf_dg_outliers_real}
\setlength{\tabcolsep}{1mm}
    \begin{tabular}{|c|cccc|cccc|}
    \hline
    \multirow{2}{*}{DAG} & \multicolumn{4}{c|}{LOF} & \multicolumn{4}{c|}{\# Outliers} \\
     & \rnd{} & \tvae{} & \ctgan{} & \gencda{} & \rnd{} & \tvae{} & \ctgan{} & \gencda{} \\
     \hline
    \abalone{} & $15.2$ & \textbf{0.11} & $0.06$ & $9.36$ & $145$ & $155$ & \textbf{152} & $108$ \\
    \old{} & $0.53$ & $0.42$ & $0.44$ & \textbf{0.35} & $66$ & $14$ & $59$ & \textbf{13} \\
    \dwd{} & $2.27$ & \textbf{0.17} & $0.70$ & $0.40$& $76$ & $24$ & $109$ & \textbf{16} \\
    \undata{} & $0.31$ & $0.45$ & \textbf{0.29} & $3.81$ &  $17$ & \textbf{1} & $28$ & $192$ \\
    \diabets{} & \textbf{0.25} & $0.45$ & $0.35$ & $0.37$ & $42$ & \textbf{1} & $16$ & $26$ \\
    \hline
    \end{tabular}
\end{table}

\begin{figure}[t]
\centering
    \includegraphics[width=.72\linewidth]{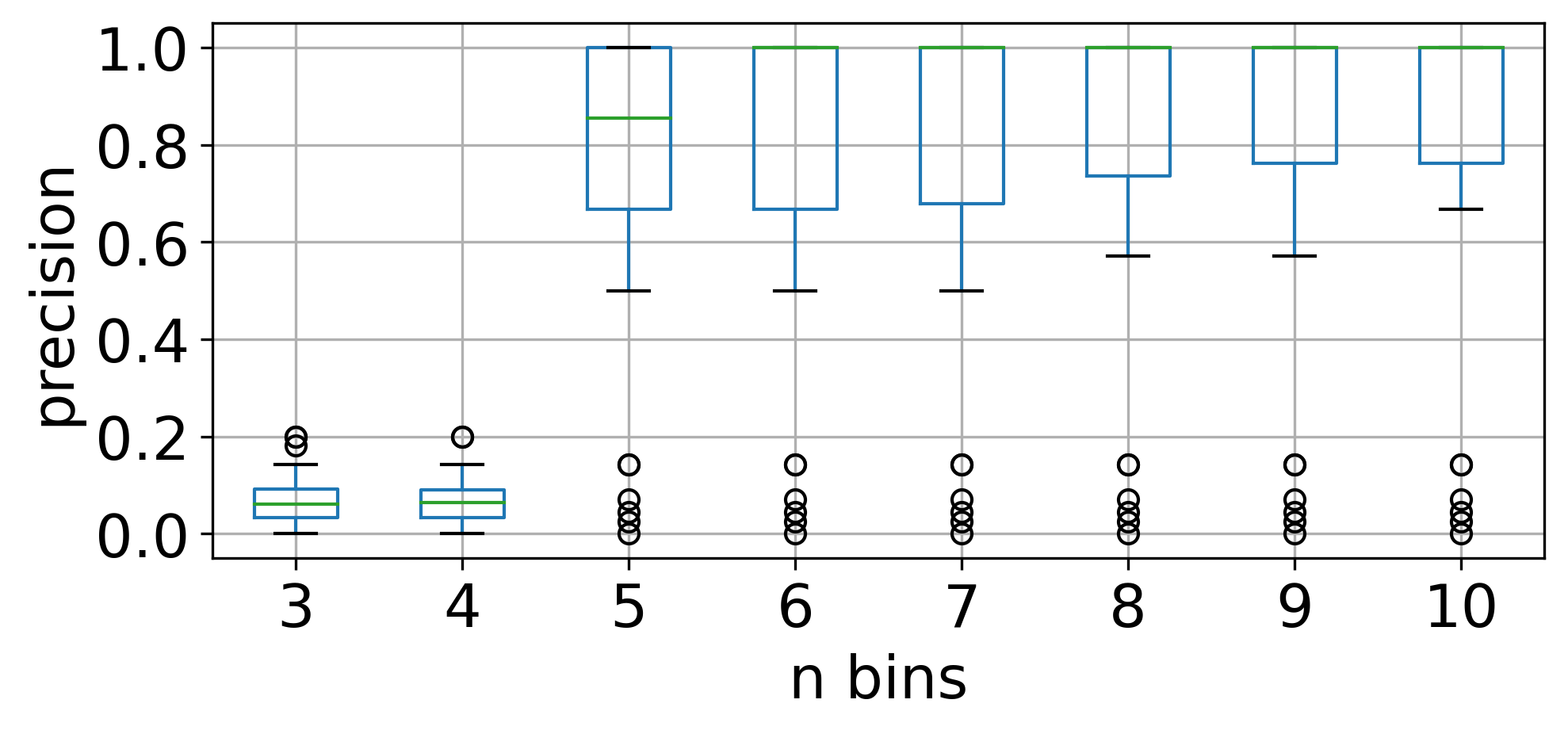}
    \includegraphics[width=.72\linewidth]{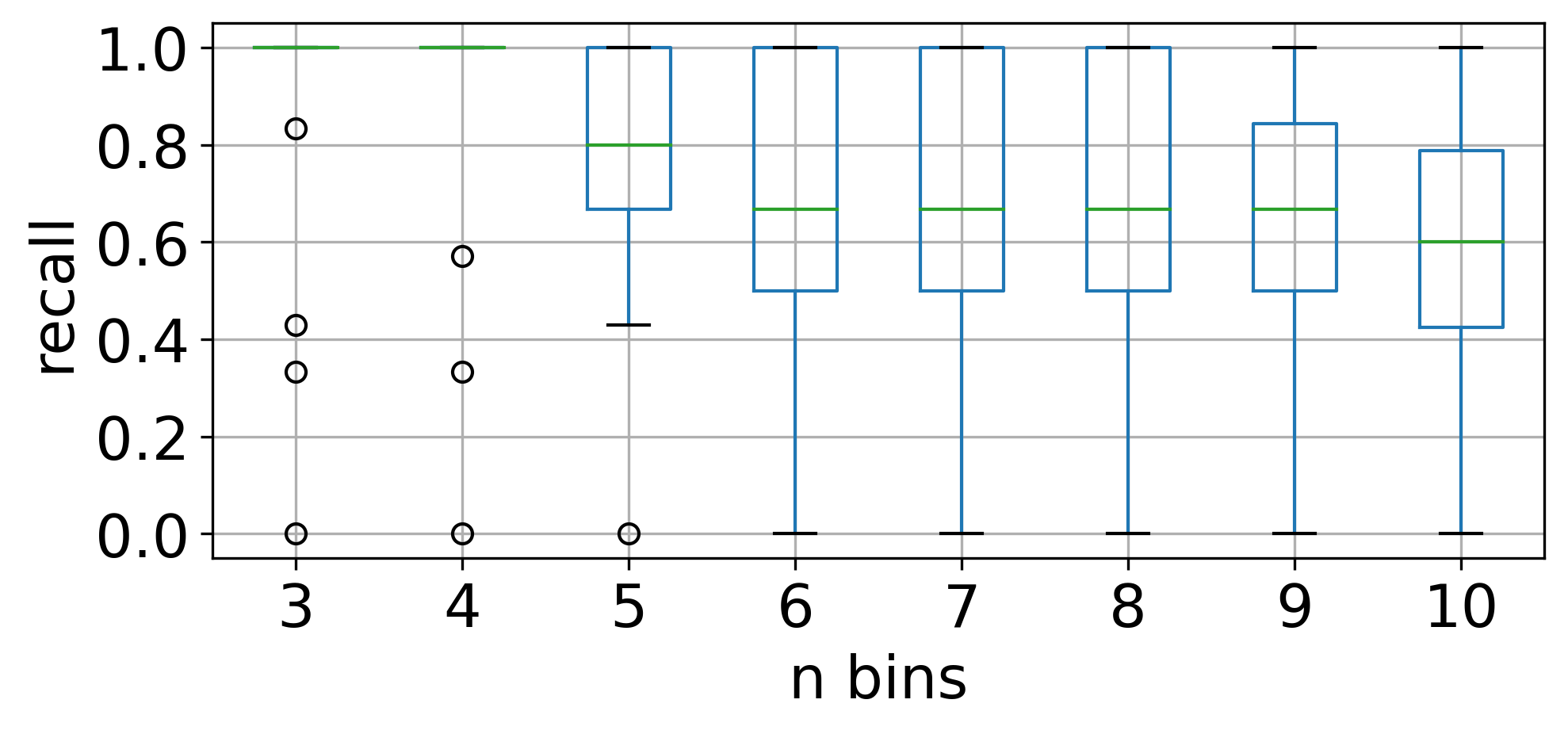} 
    \caption{Effect of $\mathit{n\_bins}$ on the performance of \ncda{}.}
    \label{fig:nbr_bins}
\end{figure}

\begin{figure}[t]
\centering
    \includegraphics[width=.72\linewidth]{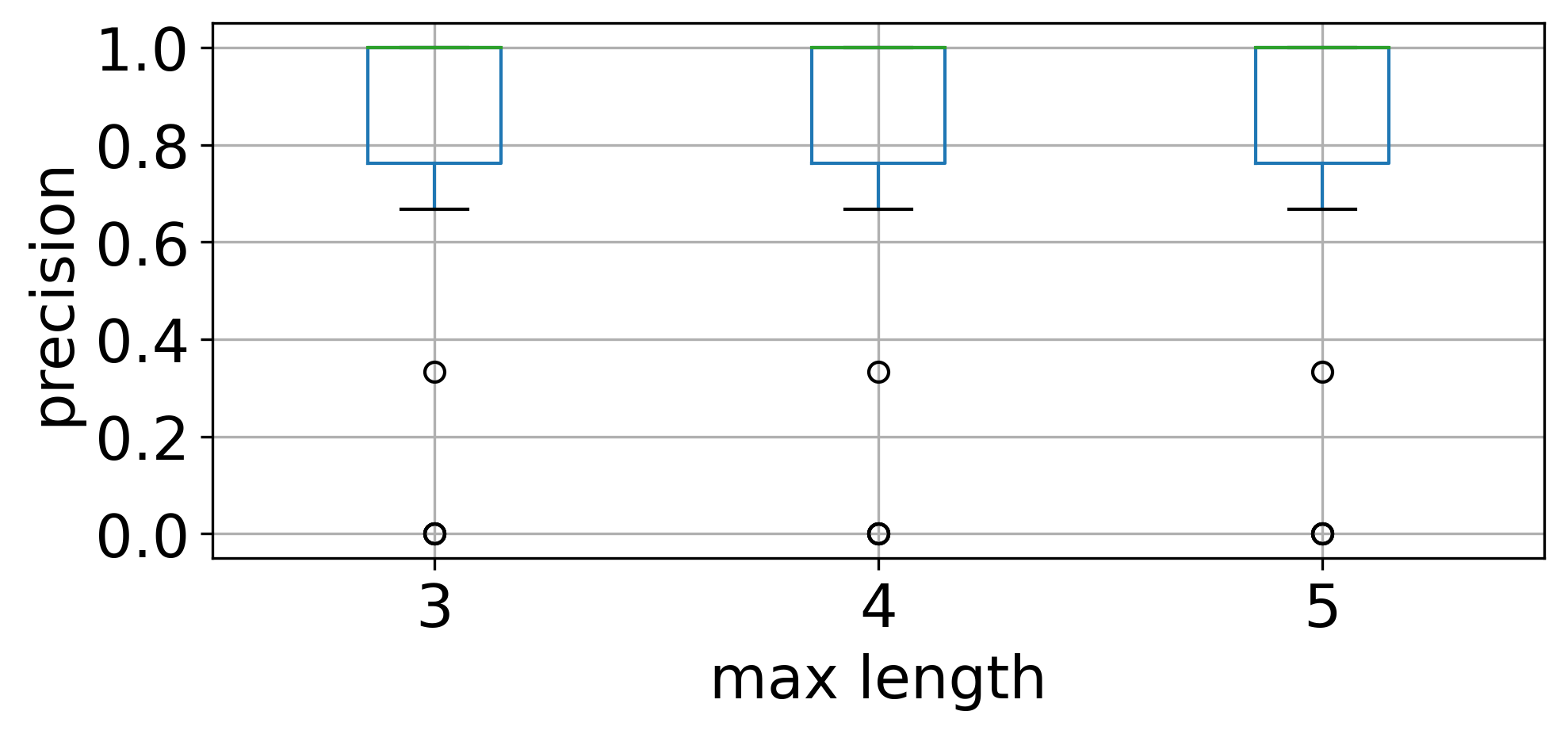}
    \caption{Effect of $\mathit{max\_len}$ on the performance of \ncda{}.}
    \label{fig:max_len}
\end{figure}

\subsection{Sensitivity Analysis}
\label{sec:exp_sensitivity_analysis}
We analyze here the parameters mainly affecting the behavior of \ncda{} and \gencda{}.
In Figure~\ref{fig:nbr_bins} we observe the boxplots of precision and recall when varying $\mathit{n\_bins} \in [3, 10]$ for 50 synthetic DAGs randomly generated.
We notice that a higher number of bins improves the precision (and the accuracy), while a lower one only fosters the recall.
Since our final objective is to discover causal relationships for data generation, we prefer to be conservative and we consider the causal structure only when we are sure that we have a causal relationship.
Therefore, in the experiments we used $\mathit{n\_bins}=10$.
In Figure~\ref{fig:max_len} we observe the boxplots of precision when varying $\mathit{max\_len} \in \{3, 4, 5\}$. 
We notice that there is no difference in performance (the same applies to accuracy and recall).
Hence we use $\mathit{max\_len}=3$ as the default value. 

We set $\mathit{min\_sup} = 0.05$ for the following reasons.
First, from preliminary experimentation emerged that with 50 synthetic DAGs if $\mathit{min\_sup} \geq 0.15$, then the procedure is not able to identify maximal itemsets with at least two items.
Second, if $\mathit{min\_sup}=0.1$ it can find maximal itemsets in 44\% of the cases, while with $\mathit{min\_sup}=0.05$ this number reaches 92\%.
With $\mathit{min\_sup}=0.1$ there is a drop in the accuracy with respect to $\mathit{min\_sup}=0.05$ since it excludes itemsets that actually correspond to causal dependencies. 
Third, considering values of $\mathit{min\_sup}$ lower than $0.05$ highly increases the chances of retrieving patterns not relevant for our task, i.e., taking into account features that are not part of the underlying causal model to the data. 
Thus, since the verification of a causal relationship is done by \ncd{} we optimize \textsc{apriori} for the task of retrieving the highest possible number of admissible maximal itemsets.
Finally, we considered different p-values thresholds $\alpha \in \{0.001, 0.01, 0.02, 0.05, 0.1\}$.
Since the impact of varying $\alpha$ is negligible, we decided to keep our procedure conservative by setting the lowest value, i.e.,
$\alpha = 0.001$.


\section{Conclusion}
\label{sec:conclusion}
We have observed how \ncda{} overcomes the limitation of \ncd{} while maintaining comparable performance.
Besides, we have shown that \gencda{} produces more realistic synthetic data than those generated with trivial baselines and it is comparable with time-consuming state-of-the-art generators.
Also, \gencda{} requires fewer instances than GANs or VAEs and also works on high dimensionality settings.
%
From an applications viewpoint, the exploitation of \gencda{} that provides insights of causal mechanisms will allow circumventing plausibility concerns creating trustful scenarios for developing ML algorithms in critical domains such as healthcare.

Several future research directions are possible.
First, we would like to stress \gencda{} with larger datasets and DAGs with more variables.
Second, we have focused on continuous datasets, but it would be interesting to extend our proposal to datasets with categorical attributes.
Third, \gencda{} is indeed a framework.
Thus, it could be interesting to evaluate how it behaves when replacing \ncd{} with other approaches, and/or the ensemble of regressors with another regressor (for instance, a deep neural network) to check if it is possible to improve the performance. 
Also, it could be interesting to test \gencda{} on other data types.
Fourth, it would be nice to investigate if there are theoretical properties related to the filtering through Apriori when searching for causal relationships.
Finally, we would like to study the impact of \gencda{} used as a generative procedure of explainability approaches such as {\scshape lime} or {\scshape lore}.

\section*{Acknowledgment}
This work is partially supported by the EU Community H2020 programme under the funding schemes: G.A. 871042 \textit{SoBigData++}, G.A. 952026 \emph{Humane-AI-Net}, G.A. 952215 \emph{TAILOR}, and the ERC-2018-ADG G.A. 834756 ``XAI''. 

\bibliographystyle{IEEEtran}
\bibliography{biblio}

\end{document}